\documentclass[a4paper,fleqn]{cas-sc}

\usepackage[authoryear,longnamesfirst]{natbib}
\usepackage{subcaption}
\usepackage{pgfplots}
\pgfplotsset{compat=1.18}

\begin{document}
\let\WriteBookmarks\relax
\shorttitle{Mosaic: Data-Free KD via MoE in Distributed Environments}
\shortauthors{J. Liu et~al.}

\title [mode = title]{Mosaic: Data-Free Knowledge Distillation via Mixture-of-Experts for Heterogeneous Distributed Environments}                      



\author[1,2]{Junming Liu}[orcid=0009-0000-9131-3245]
\author[1]{Yanting Gao}
\author[3]{Yuqi Li}
\author[2]{Siyuan Meng}
\author[2]{Yifei Sun}
\author[1]{Aoqi Wu}
\author[2]{Yirong Chen}

\author[2]{Ding Wang}
\cormark[1]

\author[4]{Shiping Wen}
\cormark[1]

\cortext[cor1]{Corresponding authors}

\affiliation[1]{organization={School of Computer Science and Technology, Tongji University}, 
            city={Shanghai}, 
            postcode={201804}, 
            country={China}}

\affiliation[2]{organization={Shanghai Artificial Intelligence Laboratory}, 
            city={Shanghai}, 
            postcode={200090}, 
            country={China}}

\affiliation[3]{organization={The City University of New York},
            city={New York},
            postcode={10017}, 
            country={USA}}

\affiliation[4]{organization={Shenzhen University of Advanced Technology}, 
            city={Shenzhen}, 
            postcode={518107}, 
            country={China}}

\begin{abstract}
  Federated Learning (FL) is a decentralized machine learning paradigm that enables clients to collaboratively train models while preserving data privacy.
  However, the coexistence of model and data heterogeneity gives rise to inconsistent representations and divergent optimization dynamics across clients, ultimately hindering robust global performance.
  To transcend these challenges, we propose \textbf{Mosaic}, a novel data-free knowledge distillation framework tailored for heterogeneous distributed environments.
  Mosaic first trains local generative models to approximate each client’s personalized distribution, enabling synthetic data generation that safeguards privacy through strict separation from real data.
  Subsequently, Mosaic forms a Mixture-of-Experts (MoE) from client models based on their specialized knowledge, and distills it into a global model using the generated data.
  To further enhance the MoE architecture, Mosaic integrates expert predictions via a lightweight meta model trained on a few representative prototypes.
  Extensive experiments on standard image and multimodal benchmarks demonstrate that Mosaic consistently outperforms state-of-the-art approaches under both model and data heterogeneity.
  The source code has been published at \href{https://github.com/Wings-Of-Disaster/Mosaic}{Github}.
\end{abstract}




\maketitle


\section{Introduction}

Amid the surge of data, deep learning has made remarkable strides in both established and emerging fields \cite{Yang_2025_A, Wang_2026_Deep}. However, in real-world scenarios, data is inherently distributed across clients, as seen in domains like autonomous driving \cite{Pan_2026_Geometry}, IoT \cite{Shi_2016_IOT}, and healthcare \cite{Li_2026_Efficient}.
Due to the high cost of data acquisition and increasingly stringent privacy regulations, aggregating data for centralized training is often impractical.
In light of these limitations, Federated Learning (FL) has emerged as a compelling paradigm \cite{Zhou_2026_Rethinking, Li_2026_Decoupling}, enabling decentralized clients to collaboratively train a shared global model while retaining their private data locally, thereby addressing critical concerns surrounding data privacy and data sovereignty.

However, the promise of FL comes with significant challenges, chief among them being the pervasive heterogeneity across real-world clients \cite{FedKDH_2025_Li, Park_2026_A}.
Specifically, the fundamental assumption of IID (Independently and Identically Distributed) data across clients often breaks down in practice, creating substantial \textbf{data heterogeneity} \cite{Threats_2025_Li, Zhang_2026_FedAATKE}.
This statistical divergence across clients gives rise to \textit{client drift} \cite{Karimireddy_2020_SCAFFOLD} in standard FL algorithms like FedAvg \cite{McMahan_2017_FedAvg}, where local models inherently reflect the disparities of private data rather than aligning toward a unified global objective, ultimately degrading generalization and performance consistency.
While numerous methods have been proposed to mitigate data heterogeneity, they often operate under the premise of model homogeneity, where all clients are required to use the same architecture as the global model \cite{Nin_2021_FedNLP}.
Yet in real-world deployments, clients often exhibit significant variation in resource availability, precluding the viability of training architecturally identical models.
This leads to \textbf{model heterogeneity} \cite{Diao_2021_HeteroFL}, rendering conventional aggregation methods infeasible and hindering effective knowledge exchange among clients.

Building upon the broader complexities posed by heterogeneity, a number of approaches have been explored to alleviate its impact.
Some methods attempt to sidestep the issue by enforcing a unified, high-capacity model, but this comes at the cost of reducing the participation of resource-limited clients who fall behind due to incomplete training \cite{Fu_2023_Client}.
Others adopt partial training (PT) strategies \cite{Alam_2022_FedRolex}, where each client trains only a subset of the global model, leading to fragmented updates and instability in global integration \cite{Luo_2023_DFRD}.
Among these, Knowledge Distillation (KD) \cite{Meng_2026_PCFedKD, Liu_2026_Learnable} emerges as a particularly effective alternative, providing a flexible means of transferring knowledge across heterogeneous clients while accommodating disparities in data and model capacity.
However, many KD-based methods rely on proxy datasets to facilitate knowledge transfer, either by aggregating small portions of client data which raises privacy concerns \cite{Li_2019_FedMD} or by employing external auxiliary datasets that closely resemble the original data distribution \cite{Shang_2022_FedIC}, limiting their applicability in real-world scenarios.
This motivates the need for Data-Free Knowledge Distillation (DFKD), which eliminates the reliance on external data while retaining the core advantages of distillation \cite{Li_2025_ICCV, Min_2026_FedSGK}.

To fully unlock the potential of DFKD, it is essential to achieve two pivotal objectives: constructing \textbf{a synthetic dataset} that enables effective knowledge transfer, and identifying \textbf{a teacher model} that encapsulates comprehensive knowledge.
Most existing DFKD approaches concentrate heavily on synthesized pseudo data that either closely approximate the global data distribution to ensure reliability \cite{Zhu_2021_FedGen, Lu_2024_FedKFD}, or generate hard samples near decision boundaries to increase diversity \cite{Zhang_2022_DENSE, Zhang_2022_FedFTG, Luo_2023_DFRD}, thereby enhancing the robustness of the distilled model and addressing its knowledge deficiencies.
Yet in practice, the widely adopted strategy in DFKD---uploading local generators to the server and naively blending their parameters---often suffers from substantial instability under Non-IID settings, where both distribution shifts and catastrophic forgetting accumulate over communication rounds, ultimately hindering consistent knowledge synthesis.
While these concerns are significant, there looms a more fundamental challenge: how to construct a generalizable and powerful teacher model.
Certain methods construct an ensemble of client models and weight them by the confidence each client exhibits in specific predictions to guide the distillation process \cite{Zhang_2022_DENSE, Lu_2024_FedKFD, Luo_2023_DFRD}, but our theoretical analysis and empirical results reveal that such schemes can yield erratic teachers that even underperform vanilla global models under severe data heterogeneity.
In summary, our goal is to obtain stable generators and a knowledgeable teacher model under heterogeneous conditions.

To confront the mentioned challenges, we propose \textbf{Mosaic}, a novel DFKD framework utilizing \textbf{M}ixture-\textbf{o}f-Expert\textbf{s} with \textbf{A}dvanced \textbf{I}ntegrated \textbf{C}\hspace{0pt}ollaboration
 for Heterogeneous Distributed Environments.
Mosaic begins by training a lightweight, label-agnostic generator on each client. Compared to conditional generators, this design notably reduces model complexity, training overhead, and communication cost, making it well-suited for distributed deployment.
To enhance the transferability of synthetic samples, we adopt an adversarial training scheme \cite{Addepalli_2020_DeGAN} where the local model acts as both a discriminator to ensure fidelity and a classifier to promote diversity.
To mitigate training difficulties on clients with limited data, we further regularize the generation process using the mean and variance of hidden representations from the initialized local model, thereby aligning the synthetic distribution with the global one.
Rather than aggregating generators uploaded by clients, which can lead to instability, Mosaic strategically ensembles them to collaboratively synthesize a diverse batch by composing unique fragments from each.
In the next stage, client models are organized into per-class experts based on label availability, thereby constituting a Mixture-of-Experts (MoE) architecture. A gating network governs the allocation of inputs to the appropriate experts, while a meta model aggregates their predictions, constructing a knowledgeable teacher.
A minimal set of prototype features is extracted from each client to further refine and enhance this ensemble.
Finally, this composite teacher is distilled into a global student model using the generator ensemble.

The main contributions of this work are summarized as follows:
First, we introduce a lightweight generator tailored for deployment in distributed environments, enhancing the fidelity and diversity of synthetic samples and improving the generation capabilities of clients with limited data by utilizing only local models.
Second, by leveraging the MoE architecture, we effectively address the robustness issues in teacher models from previous methods, with our MoE-based teacher significantly outperforming prior models.
Third, extensive experiments on six image datasets, two text datasets, and one image-text multimodal dataset demonstrate the superiority of Mosaic, showing not only improved accuracy for the global model but also significant performance gains for local models on classification tasks.

\section{Related Work}
\label{related_work}

\subsection{Heterogeneous Federated Learning}

Heterogeneous federated learning (HFL) has become an important research topic due to the decentralized nature of client environments and data distributions \cite{Enhancing_2023_Li}. In such settings, learning performance is mainly limited by data heterogeneity and model heterogeneity, as Non-IID client data degrades convergence and generalization, while differences in client model architectures or computational capabilities complicate effective aggregation \cite{Alam_2022_FedRolex}. To address data heterogeneity, several recent methods have shown promising results, including FedAF \cite{Wang_2024_FedAF}, which distills client knowledge into condensed representations to mitigate label-skew-induced drift, pFedFDA \cite{McLaughlin_2024_pFedFDA}, which formulates representation learning as a generative process to adapt global classifiers to local distributions, and PA3Fed \cite{Huang_2025_PA3Fed}, which improves training stability through period-aware aggregation. However, model heterogeneity remains a major challenge in HFL, commonly addressed through parameterized training or knowledge distillation, where FedRolex \cite{Alam_2022_FedRolex} ensures architectural compatibility via rolling sub-models and FedGen \cite{Zhu_2021_FedGen} leverages a server-side generator to guide local training. Previous studies indicate that combining parameterized training with knowledge distillation leads to more robust performance \cite{Luo_2023_DFRD}, motivating Mosaic to integrate PT with KD-based fine-tuning to improve generalization and achieve state-of-the-art results across benchmarks.

\subsection{Generator-Based Federated Learning}

Generator-based approaches have been increasingly explored to alleviate data scarcity and distribution heterogeneity in federated learning by synthesizing informative samples or feature representations \cite{Fang_2025_MSFGAN}. Recent studies generate synthetic data from shared feature statistics to improve client generalization under heterogeneous architectures \cite{Niu_2026_Bridging}, while FedPA \cite{Jiang_2025_FedPA} employs lightweight generators and prototype-guided adversarial learning to produce hard feature representations that mitigate model bias. Related directions further investigate dataset condensation techniques to construct compact yet informative synthetic datasets \cite{Li_2025_DDTime} and diffusion-based generation frameworks to enrich local data while preserving privacy \cite{Chen_2026_SLATSCOG}. These methods demonstrate the effectiveness of synthetic knowledge generation for improving model robustness and generalization. However, most approaches rely on a single generative model and do not explicitly preserve heterogeneous knowledge from different clients. Mosaic addresses this limitation through a generator ensemble that maintains client-specific knowledge and collaboratively synthesizes diverse samples for downstream distillation.

\subsection{Data-Free Knowledge Transfer in Federated Learning}

KD has been widely adopted in FL as an effective mechanism for transferring knowledge from teacher models to student models while reducing communication and model heterogeneity challenges \cite{Xie_2021_Viewpoint, Lan_2025_ICCV}. However, conventional KD often relies on access to representative training data, which may be unavailable in privacy-sensitive federated settings. DFKD addresses this limitation by enabling knowledge transfer without sharing raw data \cite{Duan_2023_Towards}. By generating synthetic data for distillation, DFKD preserves data privacy while facilitating effective knowledge transfer from global teacher models to local student models.
Notably, methods such as FedGen \cite{Zhu_2021_FedGen}, DENSE \cite{Zhang_2022_DENSE}, and FedFTG \cite{Zhang_2022_FedFTG} maintain a generator exclusively on the server, where the ensemble of client models serves as the teacher to guide its training. This design eliminates the need for generator aggregation and has demonstrated both effectiveness and scalability, motivating several subsequent methods that further explore this paradigm, including DFRD \cite{Luo_2023_DFRD} and FedKFD \cite{Lu_2024_FedKFD}.
In contrast, several methods adopt a client-side generator training strategy with subsequent aggregation on the server, such as FedCVAE-KD \cite{Heinbaugh_2023_FedCVAE}. While this design often leads to higher-quality synthetic data by leveraging client-specific distributions, it may suffer from distributional shift and catastrophic forgetting due to unstable generator aggregation. Mosaic follows the latter paradigm but mitigates its drawbacks by introducing a generator ensemble in place of a single global generator, effectively reducing instability.

\section{Proposed Method}

\begin{figure}
  \centering
  \includegraphics[width=1.0\textwidth]{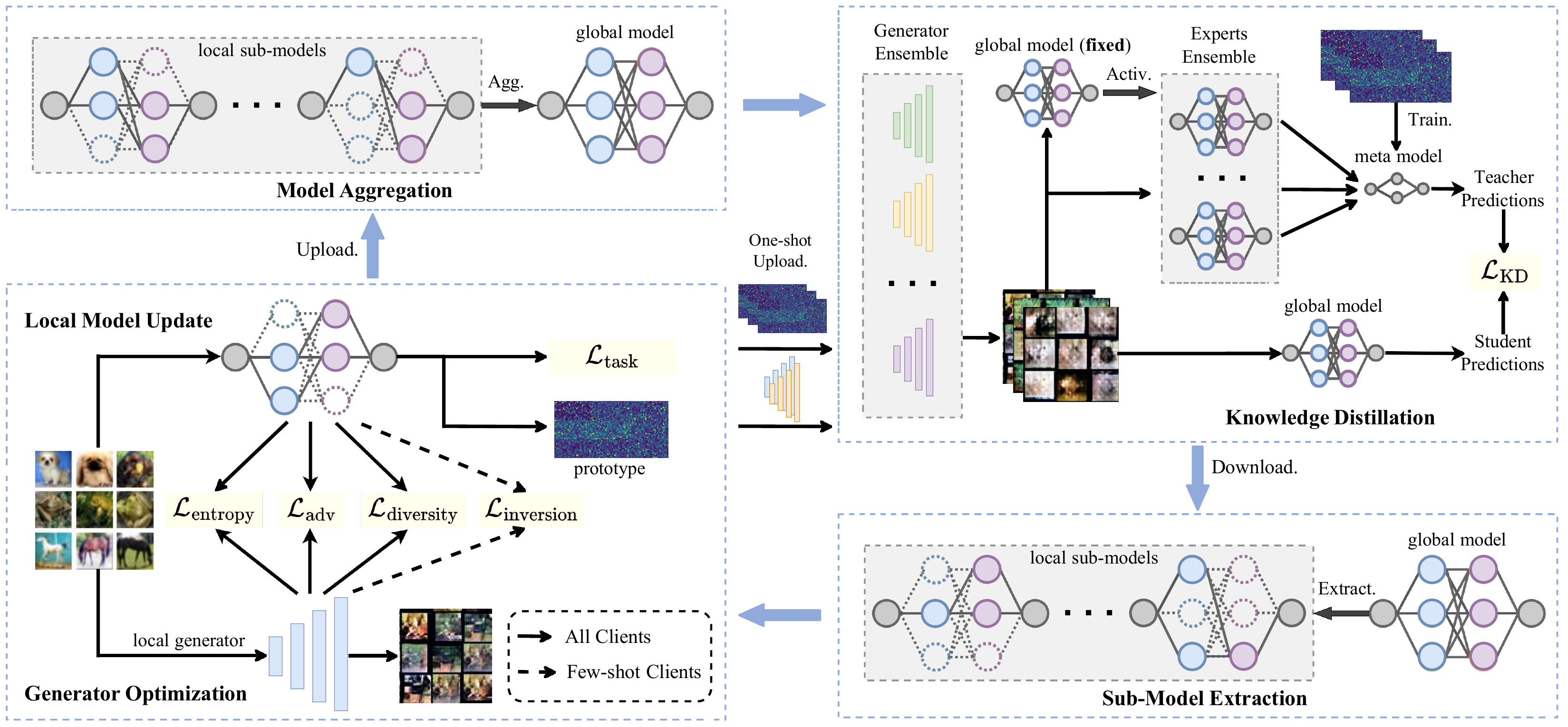}
  \caption{The full workflow for Mosaic combined with a PT-based method. Mosaic consists of four stages: \textit{local model update}, \textit{generator optimization}, \textit{model aggregation}, and \textit{knowledge distillation}. Notably, during generator optimization, the local model is updated and produces \(\mathcal{L}_{\text{adv}}\), while other losses \(\mathcal{L}_{\text{entropy}}\), \(\mathcal{L}_{\text{diversity}}\), and \(\mathcal{L}_{\text{inversion}}\) are computed using a frozen local model fixed at initialization.}
  \label{fig:workflow}
\end{figure}

In this section, we detail the proposed method Mosaic.
Mosaic is essentially a fine-tuning approach built upon PT-based methods, designed to address both model and data heterogeneity in FL environments.
It requires a preliminarily converged global model \( f \) obtained via a standard baseline method \cite{McMahan_2017_FedAvg, Alam_2022_FedRolex}, which is then distributed to initialize local models.
As illustrated in Figure~\ref{fig:workflow}, the overall workflow comprises four stages: \textit{local model update}, \textit{generator optimization}, \textit{model aggregation} and \textit{knowledge distillation}.
Notably, the process of \textit{local model update} is consistent with that in PT-based methods and is hence not elaborated here.

\subsection{Preliminaries}
\label{sec:pre}

We consider a centralized FL scenario involving a central server and $N$ clients, each possessing a private labeled dataset $\{(\mathit{X}_i, \mathit{Y}_i)\}_{i=1}^{N}$, where $\mathit{X}_i = \{x_i^b\}_{b=1}^{n_i}$ follows a local data distribution $\mathcal{D}_i$ over feature space $\mathcal{X}_i$, i.e., $x_i^b \sim \mathcal{D}_i$, and $\mathit{Y}_i = \{y_i^b\}_{b=1}^{n_i} \subseteq [C] := \{1, \dots, C\}$ denotes the corresponding ground-truth labels. Here, $C$ is the total number of classes.
The FL setting involves both data and model heterogeneity. For \textbf{data heterogeneity}, we assume that all clients share the same feature space, but their data distributions may differ typically due to label distribution skew, i.e., $\mathcal{X}_i = \mathcal{X}_j$ and $\mathcal{D}_i \neq \mathcal{D}_j,\, \forall\, i \neq j,\, i, j \in [N]$.
For \textbf{model heterogeneity}, each client $i$ maintains a local model $f_i$ with parameters $\theta_i$, and the model capacity may vary across clients, i.e., \(|\theta_i| \neq |\theta_j|, \ \exists\, i \neq j,\, i, j \in [N] \).

In PT-based methods, the capacity of client $i$ is determined by a width ratio $R_i \in (0, 1]$, indicating the proportion of neurons selected from each layer of a global model $f$ with parameters $\theta$.
During training, each client receives a sub-model with a reduced width and performs local updates. The server then aggregates updated parameters shared across clients to refine the global model. Specifically, for each parameter $\theta^t_{[l,k]}$ at round $t$, the aggregation follows \cite{Diao_2021_HeteroFL, Alam_2022_FedRolex, Luo_2023_DFRD}:
\begin{equation}
\theta^t_{[l,k]} = \frac{1}{\sum_{j \in \mathcal{S}_t} p_j} \sum_{i \in \mathcal{S}_t} p_i \, \theta^t_{i,[l,k]},
\end{equation}
where $\mathcal{S}_t \subseteq [N]$ denotes the participating clients, and $p_i$ is a weight typically proportional to the data size of client $i$, $\theta^t_{[l,k]}$ denotes the $k$-th parameter of layer $l$ of the global model, and $\theta^t_{i,[l,k]}$ denotes the parameter $\theta^t_{[l,k]}$ updated by client $i$.
Parameters not updated by any client remain unchanged.

\subsection{Generator Optimization}

In this stage, our goal is to train well-behaved generators that enhance the transferability of synthetic data, thereby improving the efficacy of KD.
Many DFKD methods adopt a conditional generator that participates in multiple rounds of communication \cite{Sun_2022_Fed2KD, Sun_2025_Fed2KD+}.
While effective, this paradigm imposes significant burdens on clients in terms of storage, computation, and communication.
Specifically, each client must continuously train and maintain a relatively large conditional generator, and transmit both the local model and the generator during every round.
Moreover, under severe data heterogeneity, we observe that the aggregated generator on the server tends to deviate from the global distribution and suffer from catastrophic forgetting.
Meanwhile, the generator deployed to clients, especially in Generative Adversarial Network (GAN)-based settings, often collapses due to mismatches with local data and discriminator, leading to training degradation \cite{Zhang_2019_GAN}.

\begin{figure}{!t}
  \centering
  \includegraphics[scale=0.55]{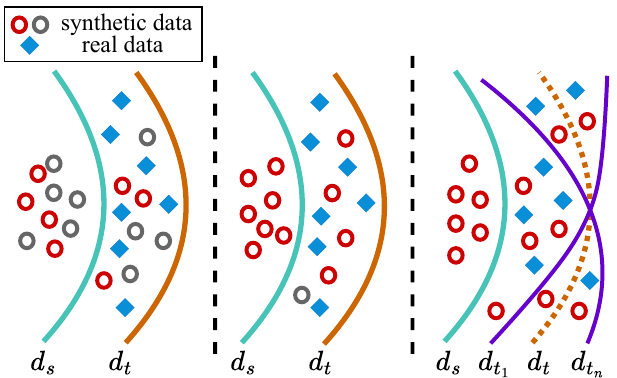}
  \caption{Visualization of synthetic data and decision boundaries of the global model \(d_s\) and teacher ensemble \(d_t\). \textit{Left}: data from the aggregated global generator, with gray regions indicating unstable coverage. \textit{Middle}: data synthesized by the generator ensemble with improved coverage. \textit{Right}: decision boundaries of the original (dashed) and MoE-based (solid) teacher ensembles.}
  \label{problem}
\end{figure}

To mitigate these issues, Mosaic employs a lightweight unconditional generator design operating under a one-shot communication protocol \cite{Heinbaugh_2023_FedCVAE}, where the generator is uploaded only once, eliminating the need for iterative synchronization.
As a result, Mosaic significantly reduces overall costs.
This naturally raises an interesting question: \textbf{Why aggregate generators when they are uploaded only once?}
Aggregating the generators can introduce instability, and the diversity of a single global generator is also limited.
Motivated by these findings, we preserve all client generators as an ensemble without aggregation. In the subsequent KD stage, this ensemble collectively synthesizes data batches, thereby avoiding distributional shifts and catastrophic forgetting induced by parameter fusion while substantially enhancing sample diversity through parallel generation.

In the following, we detail the training of the generators.
We adopt adversarial training under the GAN framework, as it offers significantly lower computational cost than Diffusion models \cite{Li_2026_Noisy} and generates higher-quality, more diverse samples than Variational Autoencoders (VAEs) \cite{Liu_2025_FedRecon}.
Additionally, we leverage the local model \( f_i \) as the discriminator on each client, avoiding the extra overhead of training a dedicated one.
To ensure \textbf{fidelity}, we utilize the basic adversarial loss \( \mathcal{L}_{\text{adv}} \) to guide the generator \( G_i \) in producing samples that align with the local data distribution \( \mathcal{D}_i \). The generator’s adversarial loss is formulated as:
\begin{equation}
\mathcal{L}_{\text{adv,real}} = \mathbb{E}_{x \sim \mathcal{D}_i}[\log (\max_k \sigma(f_i(x))_k)], \quad
\mathcal{L}_{\text{adv,fake}} = \mathbb{E}_{z \sim \mathcal{Z}}[\log(1 - \max_k \sigma(f_i(G_i(z)))_k)],
\end{equation}
where \( x \sim \mathcal{D}_i \) represents a sample drawn from \( \mathcal{D}_i \), \( z \sim \mathcal{Z} \) is a latent vector sampled from a standard Gaussian distribution, \( \sigma(\cdot) \) denotes the softmax function, and \( k \) is the class index.
Inspired by DeGAN \cite{Addepalli_2020_DeGAN}, we use\( f_i \) as an additional classifier to enhance sample confidence.
By incorporating an entropy loss \(\mathcal{L}_{\text{entropy}} \), we guide \( G_i \) to produce high-confidence samples belonging to the correct class. This entropy loss is formulated as:
\begin{equation}
\mathcal{L}_{\text{entropy}} = \mathbb{E}_{z \sim \mathcal{Z}} \left[ - \sum_{k=0}^{C} y_k \log(y_k) \right],
\end{equation}
where \( y = f_i(G_i(z)) \) is the classifier output for a generated sample, and \( y_k \) denotes the predicted probability for class \( k \), ensuring that each generated sample is confidently assigned to a class.

To further enhance the \textbf{diversity} of the generated samples, we incorporate a diversity loss \( \mathcal{L}_{\text{diversity}} \) following the same formulation as in DeGAN, which encourages \( G_i \) to produce a more varied set of samples. This diversity loss is formulated as:
\begin{equation}
\mathcal{L}_{\text{diversity}} = - \sum_{k=0}^{C} w_k \log(w_k),
\end{equation}
where \( w_k = \mathbb{E}_{z \sim \mathcal{Z}}[y_k] \) denotes the expected probability for class \( k \) by the classifier \( f_i \) over a batch of generated samples, ensuring that the overall class distribution is as uniform as possible.

However, we observe that the generators trained on clients with limited data perform poorly due to overfitting of the discriminator and mode collapse.
This is because the small dataset causes an imbalance in the adversarial training, leading the discriminator to prematurely converge and lose its guiding capability, which forces the generator to produce only a limited range of samples to deceive the discriminator.
To address this issue, we introduce the inversion loss:
\begin{equation}
\mathcal{L}_{\text{inversion}} = 
\sum_{l=1}^{L} \left\| \mu_l(G_i(z)) - \hat{\mu}_l \right\|_2
\;+\;
\sum_{l=1}^{L} \left\| \sigma_l^2(G_i(z)) - \hat{\sigma}_l^2 \right\|_2,
\end{equation}
where \( \mu_l(G_i(z)) \) and \( \sigma_l^2(G_i(z)) \) denote the batch-wise mean and variance of the feature maps at the \(l\)-th layer of the global model \(f\) given synthetic data generated by \(G_i\), and \( \hat{\mu}_l \), \( \hat{\sigma}_l^2 \) are the running mean and variance stored in the BatchNorm layers \cite{Yin_2020_DeepInversion} of \(f\), which approximate the global feature statistics accumulated during global training.

To further refine the application of the inversion loss, we introduce an empirical threshold \( \tau \), which defines the minimum sample count below which generators trained on clients with limited data are eligible to incorporate the inversion loss. While this might cause \( G_i \) to deviate from its own data distribution, the overall impact is minimal due to the limited sample size. By applying the inversion loss, we reduce the risk of overfitting and mode collapse, thereby enhancing both the quality and diversity of the generated samples.
Alternatively, $\tau$ can be set adaptively as $\tau_{\text{adaptive}} = \beta \cdot \text{median}(\{|D_i|\}_{i=1}^N)$ with $\beta \in (0,1]$ to accommodate domains with varying data scales.
The total generator loss \( \mathcal{L}_G \) is now expressed as:
\begin{equation}
\mathcal{L}_G = \mathcal{L}_{\text{adv}} + \lambda_e \mathcal{L}_{\text{entropy}} - \lambda_d \mathcal{L}_{\text{diversity}} + \lambda_i \mathcal{L}_{\text{inversion}},
\label{G_loss}
\end{equation}
where \( \lambda_e \), \( \lambda_d \), and \( \lambda_i \) are hyperparameters controlling the relative importance of the entropy, diversity, and inversion losses, respectively.
By jointly optimizing these objectives, the generator ensemble effectively reduces the instability introduced by aggregation, which manifests as deviations from the global distribution and the catastrophic forgetting of certain hard samples. This leads to improved generation quality and significantly enhanced diversity.
Since the generators capture distributional statistics rather than sample-level details, the one-shot paradigm remains effective as long as local distributions are relatively stable; the continuously updated client models further ensure that the MoE teacher reflects the latest knowledge.
Notably, the synthetic data from well-trained generators should differ visually from real data for privacy, while still capturing common knowledge from local models to align with the real data distribution for utility \cite{Luo_2023_DFRD}. 
We discuss privacy protection in detail in Section~\ref{exp}, where we elaborate on how our method addresses privacy preservation and the potential limitations that may arise.

\subsection{Model Aggregation}
\label{sec:moe}

We now describe how to construct a MoE architecture.
Our objective is to ensure that each expert specializes in a distinct subset of knowledge, with minimal redundancy or overlap across experts. Such orthogonality not only enhances task coverage but also reduces the server-side overhead of maintaining overlapping models.
To this end, we aggregate client models in a class-specific manner.
Specifically, for each class \( c \in [C] \), we construct a corresponding expert \( f_c \) by selectively combining the local models \(\{f_i\}_{i=1}^N\) according to the relative availability of class-\(c\) data:
\begin{equation}
\theta_c^{(t+1)} = \sum_{i=1}^{N} \frac{|D_{i,c}|}{\sum_{j=1}^{N} |D_{j,c}|} \, \theta_i^{(t)},
\label{eq:class_aggregation}
\end{equation}
where \( \theta_i^{(t)} \) denotes the parameters of local model \( f_i \) at round \( t \), and \( D_{i,c} = \{(x_i^b, y_i^b) \in (X_i, Y_i) \mid y_i^b = c\} \) is the subset\footnote{For clarity, as defined in Section~\ref{sec:pre}, we distinguish between the data distribution $\mathcal{D}_i$ and the empirical dataset $D_i$, where the latter is a set of samples drawn from the former (i.e., $D_i \sim \mathcal{D}_i$).} of client \( i \)'s data labeled as class \( c \).
This aggregation ensures that each expert \( f_c \) is dominated by contributions from clients who are most knowledgeable about class \( c \), thereby promoting specialization while avoiding errors introduced by uninformed contributors.
By repeating this process for all classes \( c \in [C] \), the server obtains a set of class-specific experts \( \mathcal{F} = \{f_c\}_{c=1}^C \), which together form the expert ensemble for downstream distillation or inference.

However, this design introduces a potential scalability issue when the number of classes \( C \) becomes very large. Maintaining a full set of class-specific experts incur prohibitive storage and computational costs on the server, even with a gating network that selectively activates relevant experts during inference.
Fortunately, we observe that in the large-\(C\) regimes, the expert ensemble can be substituted without significant degradation in performance.
When \( C \gg N \) and \( N \) is bounded, extreme data heterogeneity implies that each client \( i \)'s label set \( Y_i \subseteq [C] \) is nearly disjoint, i.e., \( |\left( Y_i \setminus Y_j \right) \cup \left( Y_j \setminus Y_i \right)| \gg 0, \forall\,i\neq j\in[N] \).
This suggests that certain classes are exclusive to individual clients, and the corresponding experts \( f_c \) effectively approximate the local model \( f_i \).
This motivates the consideration of directly ensembling local models, similar to the DENSE~\cite{Zhang_2022_DENSE} approach.
In contrast, when \( N \gg C \) and \( N \) is bounded, class-wise aggregation remains more effective due to the increased overlap in clients' label distributions. We empirically observe this phenomenon, confirming the relative effectiveness of this approach in such regimes.
When both $C$ and $N$ are large, the expert count can be reduced by partitioning classes into $K$ groups, each served by a single shared expert, since the gating mechanism activates only top-$k$ experts per sample regardless of the total expert count.

The gating network we used is a replica of the global model \( f \), which dictates the selection of experts within the ensemble.
Specifically, for any synthetic sample \( \hat{x} \), we identify top \( k \) classes with the highest predicted scores from \( f \), denoted as \( \text{Top}_k(f(\hat{x})) \).
The activated experts for \( \hat{x} \) are then:
\begin{equation}
\mathcal{F}(\hat{x}) = \{ f_c \in \mathcal{F} \mid c \in \text{Top}_k(f(\hat{x})) \}.
\label{eq:activated_ensemble}
\end{equation}

To enhance the robustness of the ensemble, we employ a simple Multi-Layer Perceptron (MLP) as the meta model \( M \) to weight the outputs of \( \mathcal{F}(\hat{x}) \).
Exactly, \( M \) assigns a weight \( \alpha_{c,k} \) to each expert’s output at position \( k \), and aggregates the scaled logits as:
\begin{equation}
f_{\text{meta},k}(\hat{x}) = \sum_{f_c \in \mathcal{F}(\hat{x})} \alpha_{c,k} \cdot f_c(\hat{x})_k.
\end{equation}
The final prediction is then computed as:
\begin{equation}
f_{\text{meta}}(\hat{x}) = M\left( \left[ f_c(\hat{x}) \right]_{f_c \in \mathcal{F}(\hat{x})} \right),
\end{equation}
where \( \left[ f_c(\hat{x}) \right]_{f_c \in \mathcal{F}(\hat{x})} \) denotes the concatenation of the logits from \( f_c \) in \( \mathcal{F}(\hat{x}) \).

To facilitate the training of \( M \), we extract a minimal set of prototypes from the clients \cite{Tan_2022_FedProto}.
For client \( i \), the prototype for class \( c \) is computed as the mean of the feature vectors of all samples from class \( c \) in \( D_i \), given by:
\begin{equation}
p_c^{(i)} = \frac{1}{|D_{i,c}|} \sum_{x_{i,c} \in D_{i,c}} f_i(x_{i,c}),
\label{prototype}
\end{equation}
where \( f_i(x) \) denotes the feature representation extracted by \( f_i \), and \( D_{i,c} \) denotes the subset of \( D_i \) containing samples from class \( c \).
These prototypes are based on feature averages, providing privacy protection by reducing exposure of raw data.
Furthermore, since we only extract prototypes from a few classes, the communication cost remains minimal.
We prefer prototypes over GAN-generated synthetic samples for training $M$ because prototypes carry ground-truth labels, whereas the unconditional generators produce unlabeled outputs whose pseudo-labels would depend on the very ensemble being trained, introducing circularity. In our implementation, all participating clients contribute prototypes for every class present in their local datasets.

Once the prototypes are collected, they are passed through \( f \) and \( \mathcal{F}(\hat{x}) \), where all model parameters are kept fixed. We then train \( M \) using only a standard Cross-Entropy (CE) loss:
\begin{equation}
\mathcal{L}_{\text{meta}} = \mathrm{CE}(y_{\text{meta}}, y),
\end{equation}
where \( y \) is the ground-truth label associated with the prototype.
Following these steps across several modules, we construct a teacher model that, while slightly more complex, proves to be highly robust.

\subsection{Knowledge Distillation}

Based on the previous sections, we deploy a generator ensemble \( \{ G_i \}_{i=1}^{N} \) and a MoE model \( \mathcal{F} \) on the server side.
The synthetic data produced by \( \{ G_i \}_{i=1}^{N} \) are used to optimize the student \( f \), guided by the teacher \( \mathcal{F} \).
The KD process is performed by combining soft Kullback-Leibler (KL) divergence with hard CE loss.
Therefore, the total distillation loss is formulated as:
\begin{equation}
\label{eq:kd_loss_independent_z}
\mathcal{L}_{\text{KD}}
=
\sum_{i=1}^{N}
\mathbb{E}_{z \sim \mathcal{Z}}
\Bigl[
\lambda_{\text{soft}}\,
\mathrm{KL}\,\bigl(f\bigl(G_i(z)\bigr),\,\mathcal{F}\bigl(G_i(z)\bigr)\bigr)
+
\lambda_{\text{hard}}\,
\mathrm{CE}\,\bigl(f\bigl(G_i(z)\bigr),\,Y_i\bigr)
\Bigr],
\end{equation}
where \( Y_i = \arg\max_{c} \bigl[\mathcal{F}(G_i(z))\bigr]_c \) denotes the hard label predicted by \( \mathcal{F} \) for the synthetic sample \( G_i(z) \), with \( z \) drawn from the latent space \( \mathcal{Z} = \mathcal{N}(0, 1) \).

\section{Experiments}
\label{exp}

\begin{table}[!t]
  \centering
  \caption{Top test accuracy~(\%) across $\omega\in\{0.01,0.1,1.0\}$ on FMNIST and SVHN. 
  Numbers in parentheses denote results under stronger heterogeneity settings.}
  \begin{tabular}{lccc|ccc}
    \toprule
    \multirow{2}{*}{Alg.s} 
      & \multicolumn{3}{c|}{FMNIST} 
      & \multicolumn{3}{c}{SVHN} \\
    \cmidrule(lr){2-4}\cmidrule(lr){5-7}
      & $\omega=1.0$ & $\omega=0.1$ & $\omega=0.01$
      & $\omega=1.0$ & $\omega=0.1$ & $\omega=0.01$ \\
    \midrule
    FedAvg & 
      \begin{tabular}{c}89.89{\scriptsize$\pm$0.23}\\{\scriptsize(82.80$\pm$1.59)}\end{tabular} &
      \begin{tabular}{c}82.00{\scriptsize$\pm$1.38}\\{\scriptsize(55.62$\pm$1.60)}\end{tabular} &
      \begin{tabular}{c}57.74{\scriptsize$\pm$0.89}\\{\scriptsize(35.01$\pm$1.44)}\end{tabular} &
      \begin{tabular}{c}89.01{\scriptsize$\pm$1.67}\\{\scriptsize(79.73$\pm$1.87)}\end{tabular} &
      \begin{tabular}{c}75.37{\scriptsize$\pm$1.55}\\{\scriptsize(36.70$\pm$1.41)}\end{tabular} &
      \begin{tabular}{c}37.77{\scriptsize$\pm$1.60}\\{\scriptsize(14.88$\pm$1.62)}\end{tabular} \\
    FedRS &
      \begin{tabular}{c}90.02{\scriptsize$\pm$0.31}\\{\scriptsize(85.87$\pm$0.41)}\end{tabular} &
      \begin{tabular}{c}82.99{\scriptsize$\pm$1.22}\\{\scriptsize(57.21$\pm$0.93)}\end{tabular} &
      \begin{tabular}{c}58.33{\scriptsize$\pm$2.03}\\{\scriptsize(38.63$\pm$2.43)}\end{tabular} &
      \begin{tabular}{c}90.08{\scriptsize$\pm$1.27}\\{\scriptsize(\underline{83.84$\pm$0.69})}\end{tabular} &
      \begin{tabular}{c}77.39{\scriptsize$\pm$2.33}\\{\scriptsize(39.87$\pm$1.87)}\end{tabular} &
      \begin{tabular}{c}38.59{\scriptsize$\pm$1.99}\\{\scriptsize(15.92$\pm$1.74)}\end{tabular} \\
    FedOpt &
      \begin{tabular}{c}\underline{91.96{\scriptsize$\pm$0.57}}\\{\scriptsize(85.83$\pm$0.76)}\end{tabular} &
      \begin{tabular}{c}83.40{\scriptsize$\pm$1.04}\\{\scriptsize(57.50$\pm$1.69)}\end{tabular} &
      \begin{tabular}{c}58.56{\scriptsize$\pm$1.81}\\{\scriptsize(36.66$\pm$2.69)}\end{tabular} &
      \begin{tabular}{c}91.60{\scriptsize$\pm$1.27}\\{\scriptsize(81.45$\pm$1.61)}\end{tabular} &
      \begin{tabular}{c}78.78{\scriptsize$\pm$1.49}\\{\scriptsize(38.68$\pm$1.94)}\end{tabular} &
      \begin{tabular}{c}40.56{\scriptsize$\pm$2.33}\\{\scriptsize(15.58$\pm$2.71)}\end{tabular} \\
    FedInit &
      \begin{tabular}{c}91.75{\scriptsize$\pm$0.51}\\{\scriptsize(86.18$\pm$1.17)}\end{tabular} &
      \begin{tabular}{c}83.65{\scriptsize$\pm$1.43}\\{\scriptsize(57.44$\pm$1.55)}\end{tabular} &
      \begin{tabular}{c}60.86{\scriptsize$\pm$1.70}\\{\scriptsize(39.21$\pm$2.39)}\end{tabular} &
      \begin{tabular}{c}91.55{\scriptsize$\pm$1.72}\\{\scriptsize(82.85$\pm$1.83)}\end{tabular} &
      \begin{tabular}{c}79.80{\scriptsize$\pm$2.02}\\{\scriptsize(40.23$\pm$2.72)}\end{tabular} &
      \begin{tabular}{c}43.67{\scriptsize$\pm$2.09}\\{\scriptsize(19.03$\pm$2.48)}\end{tabular} \\
    FedAF & 
      \begin{tabular}{c}91.95{\scriptsize$\pm$1.04}\\{\scriptsize(85.50$\pm$1.67)}\end{tabular} &
      \begin{tabular}{c}83.43{\scriptsize$\pm$1.79}\\{\scriptsize(58.12$\pm$1.56)}\end{tabular} &
      \begin{tabular}{c}61.41{\scriptsize$\pm$1.89}\\{\scriptsize(38.53$\pm$1.61)}\end{tabular} &
      \begin{tabular}{c}91.45{\scriptsize$\pm$1.69}\\{\scriptsize(81.76$\pm$1.79)}\end{tabular} &
      \begin{tabular}{c}79.89{\scriptsize$\pm$1.51}\\{\scriptsize(40.56$\pm$1.79)}\end{tabular} &
      \begin{tabular}{c}42.98{\scriptsize$\pm$1.81}\\{\scriptsize(18.88$\pm$1.62)}\end{tabular} \\
    pFedFDA & 
      \begin{tabular}{c}91.66{\scriptsize$\pm$1.24}\\{\scriptsize(85.30$\pm$1.31)}\end{tabular} &
      \begin{tabular}{c}83.07{\scriptsize$\pm$1.57}\\{\scriptsize(58.55$\pm$1.53)}\end{tabular} &
      \begin{tabular}{c}62.55{\scriptsize$\pm$1.43}\\{\scriptsize(40.17$\pm$1.75)}\end{tabular} &
      \begin{tabular}{c}90.94{\scriptsize$\pm$1.66}\\{\scriptsize(81.82$\pm$1.46)}\end{tabular} &
      \begin{tabular}{c}80.13{\scriptsize$\pm$1.78}\\{\scriptsize(41.47$\pm$1.85)}\end{tabular} &
      \begin{tabular}{c}44.49{\scriptsize$\pm$2.03}\\{\scriptsize(20.52$\pm$1.86)}\end{tabular} \\
    PA3Fed &
      \begin{tabular}{c}\textbf{92.91{\scriptsize$\pm$0.37}}\\{\scriptsize(\underline{86.71$\pm$1.28})}\end{tabular} &
      \begin{tabular}{c}\underline{84.48{\scriptsize$\pm$1.45}}\\{\scriptsize(\underline{58.95$\pm$1.84})}\end{tabular} &
      \begin{tabular}{c}\underline{64.89{\scriptsize$\pm$2.80}}\\{\scriptsize(\underline{42.98$\pm$3.06})}\end{tabular} &
      \begin{tabular}{c}\textbf{92.13{\scriptsize$\pm$1.84}}\\{\scriptsize(80.62$\pm$1.95)}\end{tabular} &
      \begin{tabular}{c}\underline{81.69{\scriptsize$\pm$1.89}}\\{\scriptsize(\underline{41.69$\pm$2.14})}\end{tabular} &
      \begin{tabular}{c}\underline{45.50{\scriptsize$\pm$2.36}}\\{\scriptsize(\underline{21.83$\pm$2.84})}\end{tabular} \\
    DENSE &
      \begin{tabular}{c}90.09{\scriptsize$\pm$2.25}\\{\scriptsize(83.92$\pm$3.73)}\end{tabular} &
      \begin{tabular}{c}81.37{\scriptsize$\pm$2.41}\\{\scriptsize(55.24$\pm$3.63)}\end{tabular} &
      \begin{tabular}{c}58.37{\scriptsize$\pm$2.78}\\{\scriptsize(38.20$\pm$4.74)}\end{tabular} &
      \begin{tabular}{c}89.54{\scriptsize$\pm$1.44}\\{\scriptsize(80.58$\pm$2.04)}\end{tabular} &
      \begin{tabular}{c}76.91{\scriptsize$\pm$1.26}\\{\scriptsize(37.95$\pm$2.94)}\end{tabular} &
      \begin{tabular}{c}38.07{\scriptsize$\pm$3.01}\\{\scriptsize(14.37$\pm$3.40)}\end{tabular} \\
    FedFTG &
      \begin{tabular}{c}90.36{\scriptsize$\pm$0.48}\\{\scriptsize(84.28$\pm$0.91)}\end{tabular} &
      \begin{tabular}{c}83.63{\scriptsize$\pm$1.46}\\{\scriptsize(57.51$\pm$1.71)}\end{tabular} &
      \begin{tabular}{c}58.77{\scriptsize$\pm$1.95}\\{\scriptsize(39.39$\pm$2.59)}\end{tabular} &
      \begin{tabular}{c}89.66{\scriptsize$\pm$1.45}\\{\scriptsize(79.64$\pm$1.93)}\end{tabular} &
      \begin{tabular}{c}76.64{\scriptsize$\pm$1.38}\\{\scriptsize(37.36$\pm$1.85)}\end{tabular} &
      \begin{tabular}{c}38.59{\scriptsize$\pm$2.06}\\{\scriptsize(15.50$\pm$2.38)}\end{tabular} \\
    DFRD &
      \begin{tabular}{c}91.51{\scriptsize$\pm$0.30}\\{\scriptsize(85.76$\pm$0.38)}\end{tabular} &
      \begin{tabular}{c}83.72{\scriptsize$\pm$1.46}\\{\scriptsize(58.37$\pm$1.90)}\end{tabular} &
      \begin{tabular}{c}59.48{\scriptsize$\pm$1.49}\\{\scriptsize(38.32$\pm$2.42)}\end{tabular} &
      \begin{tabular}{c}90.55{\scriptsize$\pm$1.85}\\{\scriptsize(81.61$\pm$2.30)}\end{tabular} &
      \begin{tabular}{c}79.65{\scriptsize$\pm$2.09}\\{\scriptsize(41.36$\pm$2.22)}\end{tabular} &
      \begin{tabular}{c}42.42{\scriptsize$\pm$2.66}\\{\scriptsize(18.08$\pm$2.70)}\end{tabular} \\
    FedKFD &
      \begin{tabular}{c}90.23{\scriptsize$\pm$0.27}\\{\scriptsize(84.38$\pm$1.55)}\end{tabular} &
      \begin{tabular}{c}81.80{\scriptsize$\pm$1.50}\\{\scriptsize(57.42$\pm$1.46)}\end{tabular} &
      \begin{tabular}{c}59.33{\scriptsize$\pm$1.72}\\{\scriptsize(36.94$\pm$1.66)}\end{tabular} &
      \begin{tabular}{c}90.44{\scriptsize$\pm$1.75}\\{\scriptsize(81.25$\pm$1.99)}\end{tabular} &
      \begin{tabular}{c}77.03{\scriptsize$\pm$2.19}\\{\scriptsize(38.52$\pm$2.22)}\end{tabular} &
      \begin{tabular}{c}39.48{\scriptsize$\pm$2.70}\\{\scriptsize(16.69$\pm$2.82)}\end{tabular} \\
    Mosaic &
      \begin{tabular}{c}90.42{\scriptsize$\pm$0.03}\\{\scriptsize(\textbf{90.39$\pm$0.08})}\end{tabular} &
      \begin{tabular}{c}\textbf{85.40{\scriptsize$\pm$0.61}}\\{\scriptsize(\textbf{74.22$\pm$0.55})}\end{tabular} &
      \begin{tabular}{c}\textbf{67.53{\scriptsize$\pm$0.76}}\\{\scriptsize(\textbf{56.51$\pm$1.33})}\end{tabular} &
      \begin{tabular}{c}\underline{91.79{\scriptsize$\pm$1.58}}\\{\scriptsize(\textbf{89.10$\pm$1.22})}\end{tabular} &
      \begin{tabular}{c}\textbf{82.11{\scriptsize$\pm$1.23}}\\{\scriptsize(\textbf{80.47$\pm$1.51})}\end{tabular} &
      \begin{tabular}{c}\textbf{52.28{\scriptsize$\pm$1.70}}\\{\scriptsize(\textbf{51.11$\pm$1.24})}\end{tabular} \\
    \bottomrule
  \end{tabular}
  \label{data_hetero_1}
\end{table}

\subsection{Experimental Settings}
\label{settings}

\begin{table}[!t]
  \centering
  \caption{Top test accuracy~(\%) across $\omega\in\{0.01,0.1,1.0\}$ on CIFAR-10 and CIFAR-100. 
  Numbers in parentheses denote results under stronger heterogeneity settings.}
  \begin{tabular}{lccc|ccc}
    \toprule
    \multirow{2}{*}{Alg.s} 
      & \multicolumn{3}{c|}{CIFAR-10} 
      & \multicolumn{3}{c}{CIFAR-100} \\
    \cmidrule(lr){2-4}\cmidrule(lr){5-7}
      & $\omega=1.0$ & $\omega=0.1$ & $\omega=0.01$
      & $\omega=1.0$ & $\omega=0.1$ & $\omega=0.01$ \\
    \midrule
    FedAvg & 
      \begin{tabular}{c}78.34{\scriptsize$\pm$2.19}\\{\scriptsize(61.23$\pm$1.78)}\end{tabular} &
      \begin{tabular}{c}56.32{\scriptsize$\pm$2.17}\\{\scriptsize(26.51$\pm$1.96)}\end{tabular} &
      \begin{tabular}{c}36.76{\scriptsize$\pm$2.45}\\{\scriptsize(17.17$\pm$1.92)}\end{tabular} &
      \begin{tabular}{c}65.55{\scriptsize$\pm$0.53}\\{\scriptsize(58.72$\pm$0.98)}\end{tabular} &
      \begin{tabular}{c}59.30{\scriptsize$\pm$0.53}\\{\scriptsize(41.62$\pm$0.87)}\end{tabular} &
      \begin{tabular}{c}48.72{\scriptsize$\pm$0.92}\\{\scriptsize(19.07$\pm$0.64)}\end{tabular} \\
    FedRS &
      \begin{tabular}{c}78.58{\scriptsize$\pm$1.95}\\{\scriptsize(63.37$\pm$2.10)}\end{tabular} &
      \begin{tabular}{c}57.48{\scriptsize$\pm$2.09}\\{\scriptsize(28.27$\pm$2.48)}\end{tabular} &
      \begin{tabular}{c}41.75{\scriptsize$\pm$2.07}\\{\scriptsize(18.92$\pm$2.22)}\end{tabular} &
      \begin{tabular}{c}63.32{\scriptsize$\pm$0.62}\\{\scriptsize(59.34$\pm$0.86)}\end{tabular} &
      \begin{tabular}{c}58.29{\scriptsize$\pm$0.66}\\{\scriptsize(41.45$\pm$0.90)}\end{tabular} &
      \begin{tabular}{c}50.16{\scriptsize$\pm$1.00}\\{\scriptsize(19.84$\pm$1.35)}\end{tabular} \\
    FedOpt &
      \begin{tabular}{c}81.30{\scriptsize$\pm$1.68}\\{\scriptsize(62.45$\pm$1.63)}\end{tabular} &
      \begin{tabular}{c}59.17{\scriptsize$\pm$1.83}\\{\scriptsize(27.98$\pm$1.84)}\end{tabular} &
      \begin{tabular}{c}41.86{\scriptsize$\pm$1.91}\\{\scriptsize(18.23$\pm$1.82)}\end{tabular} &
      \begin{tabular}{c}68.60{\scriptsize$\pm$0.52}\\{\scriptsize(\underline{60.91$\pm$0.89})}\end{tabular} &
      \begin{tabular}{c}61.17{\scriptsize$\pm$0.85}\\{\scriptsize(41.19$\pm$0.91)}\end{tabular} &
      \begin{tabular}{c}50.62{\scriptsize$\pm$1.06}\\{\scriptsize(20.69$\pm$1.70)}\end{tabular} \\
    FedInit &
      \begin{tabular}{c}81.18{\scriptsize$\pm$1.72}\\{\scriptsize(62.88$\pm$1.94)}\end{tabular} &
      \begin{tabular}{c}59.78{\scriptsize$\pm$1.98}\\{\scriptsize(28.37$\pm$2.50)}\end{tabular} &
      \begin{tabular}{c}42.99{\scriptsize$\pm$2.12}\\{\scriptsize(20.04$\pm$2.22)}\end{tabular} &
      \begin{tabular}{c}\underline{69.44{\scriptsize$\pm$0.68}}\\{\scriptsize(60.45$\pm$0.74)}\end{tabular} &
      \begin{tabular}{c}61.53{\scriptsize$\pm$0.53}\\{\scriptsize(42.66$\pm$0.96)}\end{tabular} &
      \begin{tabular}{c}51.27{\scriptsize$\pm$1.04}\\{\scriptsize(20.88$\pm$1.51)}\end{tabular} \\
    FedAF & 
      \begin{tabular}{c}79.81{\scriptsize$\pm$1.95}\\{\scriptsize(61.31$\pm$1.68)}\end{tabular} &
      \begin{tabular}{c}58.69{\scriptsize$\pm$2.17}\\{\scriptsize(26.44$\pm$1.71)}\end{tabular} &
      \begin{tabular}{c}40.98{\scriptsize$\pm$2.23}\\{\scriptsize(18.84$\pm$2.43)}\end{tabular} &
      \begin{tabular}{c}65.39{\scriptsize$\pm$0.77}\\{\scriptsize(58.06$\pm$1.05)}\end{tabular} &
      \begin{tabular}{c}60.28{\scriptsize$\pm$0.84}\\{\scriptsize(42.94$\pm$1.39)}\end{tabular} &
      \begin{tabular}{c}50.97{\scriptsize$\pm$1.65}\\{\scriptsize(20.06$\pm$1.83)}\end{tabular} \\
    pFedFDA & 
      \begin{tabular}{c}\underline{81.73{\scriptsize$\pm$2.12}}\\{\underline{\scriptsize(64.92$\pm$2.60)}}\end{tabular} &
      \begin{tabular}{c}60.12{\scriptsize$\pm$2.24}\\{\underline{\scriptsize(30.64$\pm$2.88)}}\end{tabular} &
      \begin{tabular}{c}41.84{\scriptsize$\pm$2.42}\\{\scriptsize(21.56$\pm$2.46)}\end{tabular} &
      \begin{tabular}{c}68.04{\scriptsize$\pm$0.77}\\{\scriptsize(59.31$\pm$1.11)}\end{tabular} &
      \begin{tabular}{c}61.23{\scriptsize$\pm$0.97}\\{\underline{\scriptsize(43.46$\pm$1.45)}}\end{tabular} &
      \begin{tabular}{c}51.72{\scriptsize$\pm$1.37}\\{\scriptsize(20.61$\pm$1.70)}\end{tabular} \\
    PA3Fed &
      \begin{tabular}{c}81.56{\scriptsize$\pm$1.91}\\{\scriptsize(63.02$\pm$2.37)}\end{tabular} &
      \begin{tabular}{c}\underline{61.50{\scriptsize$\pm$2.15}}\\{\scriptsize(29.68$\pm$2.46)}\end{tabular} &
      \begin{tabular}{c}\underline{44.19{\scriptsize$\pm$2.55}}\\{\scriptsize(\underline{23.51$\pm$2.68})}\end{tabular} &
      \begin{tabular}{c}67.83{\scriptsize$\pm$0.43}\\{\scriptsize(59.34$\pm$0.65)}\end{tabular} &
      \begin{tabular}{c}61.61{\scriptsize$\pm$0.73}\\{\scriptsize(42.37$\pm$0.90)}\end{tabular} &
      \begin{tabular}{c}\underline{52.05{\scriptsize$\pm$0.89}}\\{\scriptsize(\underline{21.95$\pm$1.34})}\end{tabular} \\
    DENSE &
      \begin{tabular}{c}79.56{\scriptsize$\pm$1.57}\\{\scriptsize(63.61$\pm$1.96)}\end{tabular} &
      \begin{tabular}{c}58.67{\scriptsize$\pm$1.78}\\{\scriptsize(28.60$\pm$1.79)}\end{tabular} &
      \begin{tabular}{c}36.38{\scriptsize$\pm$2.53}\\{\scriptsize(17.81$\pm$3.01)}\end{tabular} &
      \begin{tabular}{c}66.52{\scriptsize$\pm$0.65}\\{\scriptsize(58.97$\pm$0.99)}\end{tabular} &
      \begin{tabular}{c}60.09{\scriptsize$\pm$0.81}\\{\scriptsize(41.54$\pm$1.45)}\end{tabular} &
      \begin{tabular}{c}48.79{\scriptsize$\pm$1.77}\\{\scriptsize(18.95$\pm$2.67)}\end{tabular} \\
    FedFTG &
      \begin{tabular}{c}79.01{\scriptsize$\pm$1.26}\\{\scriptsize(61.96$\pm$1.33)}\end{tabular} &
      \begin{tabular}{c}58.24{\scriptsize$\pm$1.66}\\{\scriptsize(27.70$\pm$1.80)}\end{tabular} &
      \begin{tabular}{c}36.05{\scriptsize$\pm$2.40}\\{\scriptsize(17.63$\pm$2.30)}\end{tabular} &
      \begin{tabular}{c}65.71{\scriptsize$\pm$0.63}\\{\scriptsize(59.08$\pm$0.85)}\end{tabular} &
      \begin{tabular}{c}59.91{\scriptsize$\pm$0.75}\\{\scriptsize(41.81$\pm$0.90)}\end{tabular} &
      \begin{tabular}{c}49.37{\scriptsize$\pm$0.99}\\{\scriptsize(20.00$\pm$1.11)}\end{tabular} \\
    DFRD &
      \begin{tabular}{c}80.27{\scriptsize$\pm$1.43}\\{\scriptsize(63.00$\pm$1.58)}\end{tabular} &
      \begin{tabular}{c}59.80{\scriptsize$\pm$1.56}\\{\scriptsize(28.44$\pm$1.80)}\end{tabular} &
      \begin{tabular}{c}39.78{\scriptsize$\pm$2.41}\\{\scriptsize(18.78$\pm$2.79)}\end{tabular} &
      \begin{tabular}{c}67.55{\scriptsize$\pm$0.61}\\{\scriptsize(58.61$\pm$1.74)}\end{tabular} &
      \begin{tabular}{c}\underline{61.88{\scriptsize$\pm$0.64}}\\{\scriptsize(42.12$\pm$1.76)}\end{tabular} &
      \begin{tabular}{c}51.01{\scriptsize$\pm$1.54}\\{\scriptsize(20.02$\pm$2.36)}\end{tabular} \\
    FedKFD &
      \begin{tabular}{c}79.86{\scriptsize$\pm$2.27}\\{\scriptsize(63.08$\pm$2.70)}\end{tabular} &
      \begin{tabular}{c}57.89{\scriptsize$\pm$2.45}\\{\scriptsize(28.45$\pm$2.47)}\end{tabular} &
      \begin{tabular}{c}38.55{\scriptsize$\pm$3.18}\\{\scriptsize(18.89$\pm$3.39)}\end{tabular} &
      \begin{tabular}{c}66.90{\scriptsize$\pm$0.45}\\{\scriptsize(60.44$\pm$0.92)}\end{tabular} &
      \begin{tabular}{c}60.87{\scriptsize$\pm$0.84}\\{\scriptsize(43.45$\pm$0.81)}\end{tabular} &
      \begin{tabular}{c}50.31{\scriptsize$\pm$1.63}\\{\scriptsize(20.83$\pm$1.12)}\end{tabular} \\
    Mosaic &
      \begin{tabular}{c}\textbf{81.95{\scriptsize$\pm$1.47}}\\{\scriptsize(\textbf{76.27$\pm$1.80})}\end{tabular} &
      \begin{tabular}{c}\textbf{62.54{\scriptsize$\pm$1.46}}\\{\scriptsize(\textbf{59.77$\pm$1.77})}\end{tabular} &
      \begin{tabular}{c}\textbf{50.43{\scriptsize$\pm$1.76}}\\{\scriptsize(\textbf{47.95$\pm$1.32})}\end{tabular} &
      \begin{tabular}{c}\textbf{70.28{\scriptsize$\pm$0.98}}\\{\scriptsize(\textbf{67.62$\pm$0.69})}\end{tabular} &
      \begin{tabular}{c}\textbf{62.76{\scriptsize$\pm$0.70}}\\{\scriptsize(\textbf{57.93$\pm$0.76})}\end{tabular} &
      \begin{tabular}{c}\textbf{53.59{\scriptsize$\pm$0.90}}\\{\scriptsize(\textbf{43.92$\pm$0.99})}\end{tabular} \\
    \bottomrule
  \end{tabular}
  \label{data_hetero_2}
\end{table}

\textbf{Datasets}. We conduct experiments on six image classification datasets to comprehensively evaluate the effectiveness of our method: FMNIST \cite{Xiao_2017_FMNIST}, SVHN \cite{Netzer_2011_SVHN}, CIFAR-10, CIFAR-100 \cite{Krizhevsky_2009_CIFAR}, FOOD101 \cite{Bossard_2014_FOOD101}, and Tiny-ImageNet\footnote{http://cs231n.stanford.edu/tiny-imagenet-200.zip}. In addition, we consider NLP tasks including 20 Newsgroups \cite{Prieditis_1995_20Newsgroups} for text classification and OntoNotes \cite{Pradhan_2013_OntoNotes} for sequence tagging, as well as the multimodal dataset CrisisMMD \cite{Alam_2018_CrisisMMD} for multimodal classification.
To simulate data heterogeneity, following prior works \cite{Luo_2023_DFRD}, we partition the training data across clients using a Dirichlet distribution \( Dir(\omega) \), where a smaller \( \omega \) indicates a higher degree of data heterogeneity.

\textbf{Baselines}. We compare Mosaic against a comprehensive set of FL baselines, including FedAvg \cite{McMahan_2017_FedAvg}, FedProx \cite{Li_2020_FedProx}, FedRS \cite{Li_2021_FedRS}, FedOpt \cite{Reddi_2020_FedOpt}, FedInit \cite{Sun_2023_FedInit}, FedAF \cite{Wang_2024_FedAF}, PA3Fed \cite{Huang_2025_PA3Fed}, pFedFDA \cite{McLaughlin_2024_pFedFDA}, DENSE \cite{Zhang_2022_DENSE}, FedFTG \cite{Zhang_2022_FedFTG}, DFRD \cite{Luo_2023_DFRD} and FedKFD \cite{Lu_2024_FedKFD}.
The first eight methods (FedAvg to pFedFDA) represent state-of-the-art (SOTA) solutions that address \textbf{data heterogeneity} in FL without leveraging DFKD.
In contrast, the latter four methods (DENSE to FedKFD) are grounded in the DFKD paradigm and fully applicable in scenarios involving \textbf{model heterogeneity}.
Notably, these DFKD methods are essentially \textbf{fine-tuning} methods designed to enhance the performance of a preliminary global model \cite{Luo_2023_DFRD}.
In our experiments, this preliminary model is obtained using FedAvg \cite{McMahan_2017_FedAvg} in the case of homogeneous FL, and FedRolex \cite{Alam_2022_FedRolex} in the case of heterogeneous FL where PT methods are required.  

\begin{table}[t]
  \centering
  \caption{Top test accuracy~(\%) across $\rho\in\{5,10,40\}$ on SVHN and CIFAR-10 datasets. 
  Numbers in parentheses denote results under stronger heterogeneity settings.}
  \label{model_hetero_1}
    \begin{tabular}{l|ccc|ccc}
    \toprule
    \multirow{2}{*}{Alg.s} 
      & \multicolumn{3}{c|}{SVHN} 
      & \multicolumn{3}{c}{CIFAR-10} \\
    \cmidrule(lr){2-4}\cmidrule(lr){5-7}
      & $\rho=5$ & $\rho=10$ & $\rho=40$
      & $\rho=5$ & $\rho=10$ & $\rho=40$ \\
    \midrule
    FedRolex &
      \begin{tabular}{c}34.71{\scriptsize$\pm$13.68}\\{\scriptsize(14.20$\pm$2.94)}\end{tabular} &
      \begin{tabular}{c}23.48{\scriptsize$\pm$13.09}\\{\scriptsize(13.85$\pm$2.60)}\end{tabular} &
      \begin{tabular}{c}22.39{\scriptsize$\pm$15.28}\\{\scriptsize(13.59$\pm$2.11)}\end{tabular} &
      \begin{tabular}{c}21.11{\scriptsize$\pm$1.76}\\{\scriptsize(16.99$\pm$0.67)}\end{tabular} &
      \begin{tabular}{c}16.57{\scriptsize$\pm$4.40}\\{\scriptsize(16.12$\pm$0.90)}\end{tabular} &
      \begin{tabular}{c}14.37{\scriptsize$\pm$2.91}\\{\scriptsize(17.11$\pm$1.29)}\end{tabular} \\

    +DENSE &
      \begin{tabular}{c}36.58{\scriptsize$\pm$12.53}\\{\scriptsize(17.51$\pm$2.52)}\end{tabular} &
      \begin{tabular}{c}26.69{\scriptsize$\pm$13.11}\\{\scriptsize(14.49$\pm$1.96)}\end{tabular} &
      \begin{tabular}{c}24.34{\scriptsize$\pm$14.81}\\{\scriptsize(14.04$\pm$1.95)}\end{tabular} &
      \begin{tabular}{c}23.72{\scriptsize$\pm$5.48}\\{\scriptsize(17.16$\pm$0.67)}\end{tabular} &
      \begin{tabular}{c}19.65{\scriptsize$\pm$1.47}\\{\scriptsize(15.81$\pm$0.62)}\end{tabular} &
      \begin{tabular}{c}16.44{\scriptsize$\pm$1.89}\\{\scriptsize(17.26$\pm$1.34)}\end{tabular} \\

    +FedFTG &
      \begin{tabular}{c}38.07{\scriptsize$\pm$12.27}\\{\scriptsize(17.64$\pm$2.54)}\end{tabular} &
      \begin{tabular}{c}25.53{\scriptsize$\pm$13.84}\\{\scriptsize(14.51$\pm$2.21)}\end{tabular} &
      \begin{tabular}{c}24.06{\scriptsize$\pm$14.86}\\{\scriptsize(14.03$\pm$1.91)}\end{tabular} &
      \begin{tabular}{c}22.66{\scriptsize$\pm$5.24}\\{\scriptsize(17.16$\pm$1.03)}\end{tabular} &
      \begin{tabular}{c}17.79{\scriptsize$\pm$2.67}\\{\scriptsize(16.25$\pm$1.56)}\end{tabular} &
      \begin{tabular}{c}14.77{\scriptsize$\pm$2.03}\\{\scriptsize(17.70$\pm$1.14)}\end{tabular} \\

    +DFRD &
      \begin{tabular}{c}\textbf{46.30{\scriptsize$\pm$10.12}}\\{\scriptsize(18.44$\pm$1.34)}\end{tabular} &
      \begin{tabular}{c}34.78{\scriptsize$\pm$9.19}\\{\scriptsize(15.99$\pm$1.53)}\end{tabular} &
      \begin{tabular}{c}32.86{\scriptsize$\pm$15.54}\\{\scriptsize(15.17$\pm$0.66)}\end{tabular} &
      \begin{tabular}{c}26.68{\scriptsize$\pm$1.21}\\{\scriptsize(17.51$\pm$0.33)}\end{tabular} &
      \begin{tabular}{c}25.57{\scriptsize$\pm$1.37}\\{\scriptsize(16.74$\pm$0.72)}\end{tabular} &
      \begin{tabular}{c}19.86{\scriptsize$\pm$2.76}\\{\scriptsize(17.77$\pm$1.16)}\end{tabular} \\

    +FedKFD &
      \begin{tabular}{c}36.06{\scriptsize$\pm$3.82}\\{\scriptsize(17.34$\pm$2.97)}\end{tabular} &
      \begin{tabular}{c}25.71{\scriptsize$\pm$3.92}\\{\scriptsize(14.99$\pm$2.63)}\end{tabular} &
      \begin{tabular}{c}23.61{\scriptsize$\pm$4.43}\\{\scriptsize(14.73$\pm$2.13)}\end{tabular} &
      \begin{tabular}{c}23.32{\scriptsize$\pm$2.78}\\{\scriptsize(17.16$\pm$1.68)}\end{tabular} &
      \begin{tabular}{c}14.74{\scriptsize$\pm$3.44}\\{\scriptsize(17.28$\pm$3.91)}\end{tabular} &
      \begin{tabular}{c}18.51{\scriptsize$\pm$2.94}\\{\scriptsize(15.21$\pm$2.30)}\end{tabular} \\

    +Mosaic &
      \begin{tabular}{c}44.74{\scriptsize$\pm$3.43}\\{\scriptsize(\textbf{40.94$\pm$2.74})}\end{tabular} &
      \begin{tabular}{c}\textbf{40.61{\scriptsize$\pm$3.21}}\\{\scriptsize(\textbf{36.10$\pm$2.36})}\end{tabular} &
      \begin{tabular}{c}\textbf{37.09{\scriptsize$\pm$3.71}}\\{\scriptsize(\textbf{32.80$\pm$2.23})}\end{tabular} &
      \begin{tabular}{c}\textbf{31.60{\scriptsize$\pm$2.46}}\\{\scriptsize(\textbf{24.85$\pm$3.91})}\end{tabular} &
      \begin{tabular}{c}\textbf{28.74{\scriptsize$\pm$3.89}}\\{\scriptsize(\textbf{24.13$\pm$3.66})}\end{tabular} &
      \begin{tabular}{c}\textbf{30.37{\scriptsize$\pm$4.27}}\\{\scriptsize(\textbf{26.17$\pm$3.89})}\end{tabular} \\
    \bottomrule
    \end{tabular}
\end{table}

\textbf{Configurations}. Following the setup in DFRD, we perform experiments as follows. Unless otherwise specified, all experiments are conducted on a centralized network with $N=10$ active clients. To simulate different levels of data heterogeneity we vary the Dirichlet concentration parameter $\omega\in\{0.01,0.1,1.0\}$ and partition training data across clients according to a Dirichlet distribution $Dir(\omega)$.

For all pure image classification experiments we use ResNet-18 \cite{He_2016_ResNet} as the default backbone. To study model heterogeneity among image models we assign exponentially distributed model-capacity budgets per client
\(
R_i = \left[\tfrac{1}{2}\right]^{\min\left\{\sigma,\left\lfloor \rho\cdot\frac{i}{N}\right\rfloor\right\}}\quad(i\in[N]),
\)
with $\sigma=4$ and $\rho\in\{5,10,40\}$.
For NLP tasks we use DistilBERT \cite{Sanh_2019_DistilBERT} as the feature backbone. For multimodal classification we extract image features via ResNet and text features via DistilBERT, concatenate the modality features, and apply a lightweight fully connected classifier for final prediction.

For Mosaic we set loss weights $\lambda_e=1$, $\lambda_d=5$, $\lambda_i=10$ and distillation weights $\lambda_{\text{soft}}=0.8$, $\lambda_{\text{hard}}=0.2$, with empirical threshold $\tau=1000$. Our sensitivity analysis (Figure~\ref{fig:hyperparam_full}) confirms that these defaults generalize across all nine benchmarks without dataset-specific tuning. For new datasets, we recommend adjusting $\tau$ based on the average client data size, as clients with fewer than $\tau$ samples benefit most from the inversion loss. All baseline methods follow their default configurations. Unless otherwise noted, experiments are run with PyTorch on a single NVIDIA A100 GPU and results are averaged over three random seeds.

\textbf{Evaluation Metrics.} We assess FL performance using both local and global test accuracy. Local accuracy (L.\textit{acc}, in round brackets) is computed by evenly splitting the test set across clients and evaluating each local model individually, while global accuracy (G.\textit{acc}) is obtained by testing the server-side global model on the full test set.
In addition, we further evaluate the generated samples to examine both their quality and potential privacy risks. Specifically, we adopt the Inception Score (IS), the Fréchet Inception Distance (FID), the Silhouette Score (SS), and Pairwise Diversity (PD), which jointly measure fidelity, diversity, and the extent to which generated samples may unintentionally reveal information from the real data \cite{Metrics}.

\subsection{Evaluation on Image Classification Tasks}

We conducted in-depth analysis of the performance of various methods under different degrees of data heterogeneity on FMNIST, SVHN, CIFAR-10, and CIFAR-100, as shown in Tables~\ref{data_hetero_1} and~\ref{data_hetero_2}.
In the tables, \textbf{bold} values indicate the highest accuracy, while \underline{underlined} values denote the second-highest in each column.
It is evident that as the value of \( \omega \) decreases, all methods experience significant performance degradation.
However, Mosaic exhibits remarkable robustness under severe data heterogeneity.
This can be attributed to its stable generator ensemble and powerful MoE teacher, which together enable more effective knowledge transfer to the global model.
Moreover, Mosaic achieves remarkably high local accuracy (L.\textit{acc}), primarily because the distilled global model is further fine-tuned on each client after deployment.
Conceptually, the well-trained student model is further adapted to the local data distribution of each client, resulting in highly accurate local predictions.
On the other hand, when \( \omega \) is large and the client distributions are more uniform, the advantage of Mosaic becomes less pronounced. This is because the decision boundaries of the MoE teacher closely align with those of the global model, thereby limiting the additional knowledge that can be transferred during distillation.

We also evaluate the effects of different model heterogeneity distributions on various DFKD methods SVHN and CIFAR-10 (Table~\ref{model_hetero_1}), as well as Tiny-ImageNet and FOOD101 (Table~\ref{model_hetero_2}).
In this setting, we introduce a moderate degree of data heterogeneity by default, with $\omega = 0.1$.
For FedRolex, DENSE, FedFTG, and DFRD in Tables~\ref{model_hetero_1} and~\ref{model_hetero_2}, we adopt the results reported in the original DFRD paper \cite{Luo_2023_DFRD}, as our reproduction using the official code yielded lower performance. Both Mosaic and FedKFD are implemented on the same codebase to ensure fairness.
The model heterogeneity configuration introduces significant uncertainty, leading to highly variable results across datasets. Mosaic achieves the SOTA performance on SVHN, CIFAR-10, and FOOD101, with average improvements of 13.95\%, 12.89\%, and 12.47\% respectively.
Notably, Mosaic also exhibits substantial performance gains on Tiny-ImageNet, significantly outperforming existing baselines. As discussed in Section~\ref{sec:moe}, the MoE-based teacher provides a highly informative signal for distillation, achieving over 50\% accuracy even on this complex dataset. This strong supervisory signal, combined with our generative framework, ensures that Mosaic remains robust and effective as the task complexity and label space increase.

\textbf{Cross-Architecture Heterogeneity.} We further evaluate Mosaic under two stronger forms of model heterogeneity: \textit{intra-architecture scaling}, where clients use ResNet models of varying depths (ResNet-8 to ResNet-101) according to their local data volume, and \textit{cross-architecture}, where half of the clients adopt CNN-based models (ResNet) and the other half employ Transformer-based models (MobileViT \cite{Mehta_2022_MobileViT}). Local models are integrated via Mosaic into a MoE, distilled into global models, and redistributed. For the cross-architecture scenario, since heterogeneous models cannot be directly aggregated, we average the final logits of the global ResNet and the global MobileViT. As shown in Tables~\ref{tab:cross_architecture_full} and~\ref{tab:cross_architecture}, Mosaic consistently improves performance across all three datasets and architectures, demonstrating robustness in HFL.

\begin{table}[t]
  \centering
  \caption{Top test accuracy~(\%) across $\rho\in\{5,10,40\}$ on two datasets. 
  Numbers in parentheses denote results under stronger heterogeneity settings.}
  \label{model_hetero_2}
    \begin{tabular}{l|ccc|ccc}
    \toprule
    \multirow{2}{*}{Alg.s} 
      & \multicolumn{3}{c|}{Tiny-ImageNet} 
      & \multicolumn{3}{c}{FOOD101} \\
    \cmidrule(lr){2-4}\cmidrule(lr){5-7}
      & $\rho=5$ & $\rho=10$ & $\rho=40$
      & $\rho=5$ & $\rho=10$ & $\rho=40$ \\
    \midrule
    FedRolex &
      \begin{tabular}{c}9.29{\scriptsize$\pm$0.32}\\{\scriptsize(5.33$\pm$0.21)}\end{tabular} &
      \begin{tabular}{c}5.55{\scriptsize$\pm$0.40}\\{\scriptsize(2.73$\pm$0.09)}\end{tabular} &
      \begin{tabular}{c}2.50{\scriptsize$\pm$0.33}\\{\scriptsize(1.81$\pm$0.29)}\end{tabular} &
      \begin{tabular}{c}10.27{\scriptsize$\pm$1.33}\\{\scriptsize(6.86$\pm$0.12)}\end{tabular} &
      \begin{tabular}{c}5.14{\scriptsize$\pm$3.41}\\{\scriptsize(3.37$\pm$1.14)}\end{tabular} &
      \begin{tabular}{c}3.22{\scriptsize$\pm$0.52}\\{\scriptsize(2.95$\pm$0.17)}\end{tabular} \\

    +DENSE &
      \begin{tabular}{c}9.33{\scriptsize$\pm$0.06}\\{\scriptsize(5.16$\pm$0.18)}\end{tabular} &
      \begin{tabular}{c}5.40{\scriptsize$\pm$0.40}\\{\scriptsize(2.76$\pm$0.01)}\end{tabular} &
      \begin{tabular}{c}2.40{\scriptsize$\pm$0.20}\\{\scriptsize(1.82$\pm$0.33)}\end{tabular} &
      \begin{tabular}{c}10.83{\scriptsize$\pm$0.78}\\{\scriptsize(6.95$\pm$0.26)}\end{tabular} &
      \begin{tabular}{c}7.54{\scriptsize$\pm$0.46}\\{\scriptsize(3.86$\pm$0.44)}\end{tabular} &
      \begin{tabular}{c}3.07{\scriptsize$\pm$0.53}\\{\scriptsize(2.99$\pm$0.11)}\end{tabular} \\

    +FedFTG &
      \begin{tabular}{c}9.36{\scriptsize$\pm$0.23}\\{\scriptsize(5.18$\pm$0.18)}\end{tabular} &
      \begin{tabular}{c}5.68{\scriptsize$\pm$0.34}\\{\scriptsize(2.75$\pm$0.05)}\end{tabular} &
      \begin{tabular}{c}2.43{\scriptsize$\pm$0.12}\\{\scriptsize(1.81$\pm$0.23)}\end{tabular} &
      \begin{tabular}{c}10.66{\scriptsize$\pm$0.79}\\{\scriptsize(6.85$\pm$0.11)}\end{tabular} &
      \begin{tabular}{c}8.13{\scriptsize$\pm$0.82}\\{\scriptsize(3.75$\pm$0.48)}\end{tabular} &
      \begin{tabular}{c}3.06{\scriptsize$\pm$0.48}\\{\scriptsize(2.92$\pm$0.19)}\end{tabular} \\

    +DFRD &
      \begin{tabular}{c}10.93{\scriptsize$\pm$0.05}\\{\scriptsize(6.11$\pm$0.02)}\end{tabular} &
      \begin{tabular}{c}6.80{\scriptsize$\pm$0.11}\\{\scriptsize(3.35$\pm$0.04)}\end{tabular} &
      \begin{tabular}{c}2.68{\scriptsize$\pm$0.19}\\{\scriptsize(2.22$\pm$0.08)}\end{tabular} &
      \begin{tabular}{c}12.70{\scriptsize$\pm$0.79}\\{\scriptsize(8.02$\pm$0.08)}\end{tabular} &
      \begin{tabular}{c}10.58{\scriptsize$\pm$0.29}\\{\scriptsize(4.86$\pm$0.16)}\end{tabular} &
      \begin{tabular}{c}3.59{\scriptsize$\pm$0.05}\\{\scriptsize(3.34$\pm$0.15)}\end{tabular} \\

    +FedKFD &
      \begin{tabular}{c}9.38{\scriptsize$\pm$0.80}\\{\scriptsize(5.53$\pm$0.90)}\end{tabular} &
      \begin{tabular}{c}5.61{\scriptsize$\pm$0.75}\\{\scriptsize(2.76$\pm$0.78)}\end{tabular} &
      \begin{tabular}{c}5.53{\scriptsize$\pm$0.56}\\{\scriptsize(4.83$\pm$0.48)}\end{tabular} &
      \begin{tabular}{c}12.75{\scriptsize$\pm$0.69}\\{\scriptsize(7.36$\pm$0.64)}\end{tabular} &
      \begin{tabular}{c}6.57{\scriptsize$\pm$0.44}\\{\scriptsize(4.06$\pm$0.75)}\end{tabular} &
      \begin{tabular}{c}3.25{\scriptsize$\pm$0.53}\\{\scriptsize(2.98$\pm$0.18)}\end{tabular} \\

    +Mosaic &
      \begin{tabular}{c}\textbf{35.39{\scriptsize$\pm$0.31}}\\{\scriptsize(\textbf{21.64$\pm$0.52})}\end{tabular} &
      \begin{tabular}{c}\textbf{31.48{\scriptsize$\pm$0.46}}\\{\scriptsize(\textbf{16.65$\pm$0.57})}\end{tabular} &
      \begin{tabular}{c}\textbf{24.72{\scriptsize$\pm$0.74}}\\{\scriptsize(\textbf{11.45$\pm$0.33})}\end{tabular} &
      \begin{tabular}{c}\textbf{17.97{\scriptsize$\pm$0.09}}\\{\scriptsize(\textbf{17.88$\pm$0.16})}\end{tabular} &
      \begin{tabular}{c}\textbf{18.98{\scriptsize$\pm$0.10}}\\{\scriptsize(\textbf{18.78$\pm$0.08})}\end{tabular} &
      \begin{tabular}{c}\textbf{19.09{\scriptsize$\pm$0.05}}\\{\scriptsize(\textbf{18.64$\pm$0.14})}\end{tabular} \\
    \bottomrule
    \end{tabular}
\end{table}

\begin{table*}[t]
  \centering
  \small
  \setlength{\tabcolsep}{1.0pt}
  \caption{Top test accuracy~(\%) across $\omega\in\{0.01,0.1,1.0\}$ on three datasets with intra-architecture differences.}
    \begin{tabular}{lccc|ccc|ccc}
    \toprule
    \multirow{2}{*}{Alg.}
      & \multicolumn{3}{c|}{SVHN}
      & \multicolumn{3}{c|}{CIFAR-10}
      & \multicolumn{3}{c}{CIFAR-100} \\
    \cmidrule(lr){2-4}\cmidrule(lr){5-7}\cmidrule(lr){8-10}
      & $\omega=1.0$ & $\omega=0.1$ & $\omega=0.01$
      & $\omega=1.0$ & $\omega=0.1$ & $\omega=0.01$
      & $\omega=1.0$ & $\omega=0.1$ & $\omega=0.01$ \\
    \midrule
    FedAvg &
      \begin{tabular}{c}70.72{\scriptsize$\pm$2.01}\\{\scriptsize(38.98$\pm$1.98)}\end{tabular} &
      \begin{tabular}{c}62.22{\scriptsize$\pm$1.98}\\{\scriptsize(33.38$\pm$2.10)}\end{tabular} &
      \begin{tabular}{c}25.80{\scriptsize$\pm$2.56}\\{\scriptsize(12.95$\pm$2.50)}\end{tabular} &
      \begin{tabular}{c}62.70{\scriptsize$\pm$2.18}\\{\scriptsize(27.99$\pm$1.91)}\end{tabular} &
      \begin{tabular}{c}44.54{\scriptsize$\pm$1.74}\\{\scriptsize(20.96$\pm$2.07)}\end{tabular} &
      \begin{tabular}{c}26.70{\scriptsize$\pm$2.01}\\{\scriptsize(12.65$\pm$2.78)}\end{tabular} &
      \begin{tabular}{c}44.58{\scriptsize$\pm$2.45}\\{\scriptsize(21.99$\pm$2.07)}\end{tabular} &
      \begin{tabular}{c}38.03{\scriptsize$\pm$1.80}\\{\scriptsize(20.47$\pm$2.16)}\end{tabular} &
      \begin{tabular}{c}30.19{\scriptsize$\pm$2.41}\\{\scriptsize(15.13$\pm$2.17)}\end{tabular} \\
    Mosaic &
      \begin{tabular}{c}\textbf{84.89{\scriptsize$\pm$1.46}}\\{\scriptsize\textbf{(84.59$\pm$1.67)}}\end{tabular} &
      \begin{tabular}{c}\textbf{80.53{\scriptsize$\pm$1.82}}\\{\scriptsize\textbf{(76.29$\pm$2.33)}}\end{tabular} &
      \begin{tabular}{c}\textbf{47.69{\scriptsize$\pm$1.87}}\\{\scriptsize\textbf{(43.67$\pm$2.41)}}\end{tabular} &
      \begin{tabular}{c}\textbf{74.50{\scriptsize$\pm$1.66}}\\{\scriptsize\textbf{(72.29$\pm$1.82)}}\end{tabular} &
      \begin{tabular}{c}\textbf{62.21{\scriptsize$\pm$1.85}}\\{\scriptsize\textbf{(60.44$\pm$1.86)}}\end{tabular} &
      \begin{tabular}{c}\textbf{47.61{\scriptsize$\pm$1.96}}\\{\scriptsize\textbf{(44.89$\pm$1.66)}}\end{tabular} &
      \begin{tabular}{c}\textbf{61.49{\scriptsize$\pm$1.23}}\\{\scriptsize\textbf{(60.71$\pm$1.92)}}\end{tabular} &
      \begin{tabular}{c}\textbf{60.53{\scriptsize$\pm$1.58}}\\{\scriptsize\textbf{(57.05$\pm$1.78)}}\end{tabular} &
      \begin{tabular}{c}\textbf{50.65{\scriptsize$\pm$1.15}}\\{\scriptsize\textbf{(47.73$\pm$1.80)}}\end{tabular} \\
    \bottomrule
    \end{tabular}
  \label{tab:cross_architecture_full}
\end{table*}

\begin{table*}[t]
  \centering
  \small
  \setlength{\tabcolsep}{1.0pt}
  \caption{Top test accuracy~(\%) across $\omega\in\{0.01,0.1,1.0\}$ on three datasets with cross-architecture (ResNet vs.\ MobileViT) differences.}
    \begin{tabular}{lccc|ccc|ccc}
    \toprule
    \multirow{2}{*}{Alg.}
      & \multicolumn{3}{c|}{SVHN}
      & \multicolumn{3}{c|}{CIFAR-10}
      & \multicolumn{3}{c}{CIFAR-100} \\
    \cmidrule(lr){2-4}\cmidrule(lr){5-7}\cmidrule(lr){8-10}
      & $\omega=1.0$ & $\omega=0.1$ & $\omega=0.01$
      & $\omega=1.0$ & $\omega=0.1$ & $\omega=0.01$
      & $\omega=1.0$ & $\omega=0.1$ & $\omega=0.01$ \\
    \midrule
    FedAvg &
      \begin{tabular}{c}65.01{\scriptsize$\pm$4.04}\\{\scriptsize(50.75$\pm$3.56)}\end{tabular} &
      \begin{tabular}{c}55.32{\scriptsize$\pm$2.53}\\{\scriptsize(30.52$\pm$2.97)}\end{tabular} &
      \begin{tabular}{c}22.04{\scriptsize$\pm$2.68}\\{\scriptsize(11.33$\pm$2.44)}\end{tabular} &
      \begin{tabular}{c}58.06{\scriptsize$\pm$1.95}\\{\scriptsize(25.89$\pm$2.29)}\end{tabular} &
      \begin{tabular}{c}33.69{\scriptsize$\pm$2.61}\\{\scriptsize(15.53$\pm$2.62)}\end{tabular} &
      \begin{tabular}{c}23.82{\scriptsize$\pm$2.43}\\{\scriptsize(12.26$\pm$2.25)}\end{tabular} &
      \begin{tabular}{c}48.09{\scriptsize$\pm$1.69}\\{\scriptsize(20.17$\pm$1.97)}\end{tabular} &
      \begin{tabular}{c}26.33{\scriptsize$\pm$2.20}\\{\scriptsize(13.76$\pm$2.26)}\end{tabular} &
      \begin{tabular}{c}22.95{\scriptsize$\pm$2.58}\\{\scriptsize(12.40$\pm$2.10)}\end{tabular} \\
    Mosaic &
      \begin{tabular}{c}\textbf{85.39{\scriptsize$\pm$2.32}}\\{\scriptsize\textbf{(83.68$\pm$2.42)}}\end{tabular} &
      \begin{tabular}{c}\textbf{80.20{\scriptsize$\pm$1.97}}\\{\scriptsize\textbf{(78.70$\pm$2.32)}}\end{tabular} &
      \begin{tabular}{c}\textbf{51.00{\scriptsize$\pm$2.07}}\\{\scriptsize\textbf{(50.00$\pm$1.86)}}\end{tabular} &
      \begin{tabular}{c}\textbf{75.22{\scriptsize$\pm$1.73}}\\{\scriptsize\textbf{(73.01$\pm$1.82)}}\end{tabular} &
      \begin{tabular}{c}\textbf{60.97{\scriptsize$\pm$2.42}}\\{\scriptsize\textbf{(57.21$\pm$2.14)}}\end{tabular} &
      \begin{tabular}{c}\textbf{50.47{\scriptsize$\pm$2.31}}\\{\scriptsize\textbf{(45.99$\pm$2.60)}}\end{tabular} &
      \begin{tabular}{c}\textbf{61.82{\scriptsize$\pm$2.09}}\\{\scriptsize\textbf{(60.68$\pm$1.86)}}\end{tabular} &
      \begin{tabular}{c}\textbf{55.49{\scriptsize$\pm$2.74}}\\{\scriptsize\textbf{(50.62$\pm$2.48)}}\end{tabular} &
      \begin{tabular}{c}\textbf{45.81{\scriptsize$\pm$2.42}}\\{\scriptsize\textbf{(42.88$\pm$2.50)}}\end{tabular} \\
    \bottomrule
    \end{tabular}
  \label{tab:cross_architecture}
\end{table*}

\begin{table}[t]
\centering
\caption{Performance comparison on text and multimodal tasks under different heterogeneity levels.}

\begin{tabular}{l|cc|cc|cc}
\toprule
\multirow{2}{*}{Method} & \multicolumn{2}{c|}{20 Newsgroups} & \multicolumn{2}{c|}{OntoNotes} & \multicolumn{2}{c}{CrisisMMD} \\
\cmidrule(lr){2-3}\cmidrule(lr){4-5}\cmidrule(lr){6-7}
& $\omega=0.1$ & $\omega=0.01$ & $\omega=0.1$ & $\omega=0.01$ & $\omega=0.1$ & $\omega=0.01$ \\
\midrule
FedAvg & 
\begin{tabular}{c}44.95{\scriptsize$\pm$1.26}\\{\scriptsize(26.32$\pm$2.09)}\end{tabular} & 
\begin{tabular}{c}39.81{\scriptsize$\pm$1.62}\\{\scriptsize(19.75$\pm$1.75)}\end{tabular} & 
\begin{tabular}{c}50.31{\scriptsize$\pm$1.40}\\{\scriptsize(31.77$\pm$1.60)}\end{tabular} & 
\begin{tabular}{c}43.12{\scriptsize$\pm$1.88}\\{\scriptsize(28.96$\pm$2.08)}\end{tabular} &
\begin{tabular}{c}25.30{\scriptsize$\pm$2.06}\\{\scriptsize(12.48$\pm$2.25)}\end{tabular} & 
\begin{tabular}{c}10.51{\scriptsize$\pm$2.99}\\{\scriptsize(8.06$\pm$1.89)}\end{tabular} \\

FedProx &
\begin{tabular}{c}46.10{\scriptsize$\pm$1.53}\\{\scriptsize(27.60$\pm$2.05)}\end{tabular} & 
\begin{tabular}{c}41.25{\scriptsize$\pm$1.96}\\{\scriptsize(20.95$\pm$1.70)}\end{tabular} & 
\begin{tabular}{c}52.10{\scriptsize$\pm$1.87}\\{\scriptsize(33.00$\pm$1.96)}\end{tabular} & 
\begin{tabular}{c}44.55{\scriptsize$\pm$1.85}\\{\scriptsize(29.85$\pm$1.75)}\end{tabular} &
\begin{tabular}{c}26.70{\scriptsize$\pm$1.96}\\{\scriptsize(13.30$\pm$1.95)}\end{tabular} & 
\begin{tabular}{c}10.90{\scriptsize$\pm$2.48}\\{\scriptsize(9.30$\pm$2.01)}\end{tabular} \\

PA3Fed & 
\begin{tabular}{c}48.50{\scriptsize$\pm$0.39}\\{\scriptsize(28.90$\pm$1.47)}\end{tabular} & 
\begin{tabular}{c}42.50{\scriptsize$\pm$1.42}\\{\scriptsize(21.30$\pm$1.76)}\end{tabular} & 
\begin{tabular}{c}54.30{\scriptsize$\pm$1.25}\\{\scriptsize(35.00$\pm$1.82)}\end{tabular} & 
\begin{tabular}{c}46.80{\scriptsize$\pm$1.93}\\{\scriptsize(30.50$\pm$2.24)}\end{tabular} &
\begin{tabular}{c}28.80{\scriptsize$\pm$1.87}\\{\scriptsize(14.50$\pm$2.24)}\end{tabular} & 
\begin{tabular}{c}13.50{\scriptsize$\pm$2.32}\\{\scriptsize(10.20$\pm$2.68)}\end{tabular} \\

Mosaic & 
\begin{tabular}{c}\textbf{49.62{\scriptsize$\pm$1.56}}\\{\scriptsize(\textbf{45.27$\pm$1.99})}\end{tabular} & 
\begin{tabular}{c}\textbf{44.55{\scriptsize$\pm$1.62}}\\{\scriptsize(\textbf{39.89$\pm$2.27})}\end{tabular} & 
\begin{tabular}{c}\textbf{56.60{\scriptsize$\pm$1.30}}\\{\scriptsize(\textbf{52.74$\pm$1.61})}\end{tabular} & 
\begin{tabular}{c}\textbf{49.10{\scriptsize$\pm$1.77}}\\{\scriptsize(\textbf{41.64$\pm$1.73})}\end{tabular} &
\begin{tabular}{c}\textbf{31.20{\scriptsize$\pm$1.73}}\\{\scriptsize(\textbf{29.56$\pm$2.55})}\end{tabular} & 
\begin{tabular}{c}\textbf{20.70{\scriptsize$\pm$2.37}}\\{\scriptsize(\textbf{15.74$\pm$2.04})}\end{tabular} \\
\bottomrule
\end{tabular}
\label{tab:nlp}
\end{table}

\subsection{Evaluation on NLP and Multimodal Tasks}

We evaluate Mosaic on various NLP tasks, with the results summarized in Table \ref{tab:nlp}. It can be observed that Mosaic achieves SOTA performance, consistently surpassing the strong baseline PA3Fed. This success indicates that our prototype-based mechanism is effective not only for image data—which typically exhibits well-defined clusters, but also for textual data, where the feature space is often more complex and less amenable to distinct clustering.

Notably, although we only extract a minimal set of feature prototypes that may not exhaustively cover the entire distribution of the textual dataset, these representative "means" provide sufficient guidance for the meta model $M$ to learn robust decision boundaries. This demonstrates that even with partial coverage of the latent feature space, our approach effectively captures the core semantics of the data, maintaining high performance and robustness across different modalities.

\subsection{Ablation Study}

In this section, we conduct a thorough ablation study of the core modules in Mosaic under consistent experimental settings on SVHN, CIFAR-10, and FOOD101.
Our evaluation focuses on the scenario where both data and model heterogeneity are present, with parameters set to $w = 0.1$ and $\rho = 10$.

\begin{table}[t]
  \centering
  \caption{Test accuracy~(\%) comparison among different transferability constraints over SVHN and CIFAR-10 and FOOD101.}

    \begin{tabular}{l|c|c|c}
    \toprule
    G. L. & \multicolumn{1}{c|}{SVHN} & \multicolumn{1}{c|}{CIFAR-10} & \multicolumn{1}{c}{FOOD101} \\
    \midrule
    $\mathcal{L}_\text{adv}$     & 36.77{\scriptsize $\pm$3.22} (29.67{\scriptsize $\pm$3.03}) & 23.95{\scriptsize $\pm$2.37} (19.61{\scriptsize $\pm$2.69}) & 17.45{\scriptsize $\pm$0.07} (16.99{\scriptsize $\pm$0.11}) \\
    \midrule
    +$\mathcal{L}_\text{entropy}$     & 37.46{\scriptsize $\pm$3.54} (31.35{\scriptsize $\pm$3.81}) & 24.80{\scriptsize $\pm$3.11} (19.85{\scriptsize $\pm$3.96}) & 17.85{\scriptsize $\pm$0.06} (16.84{\scriptsize $\pm$0.17}) \\
    \midrule
    +$\mathcal{L}_\text{diversity}$     & 37.99{\scriptsize $\pm$3.40} (31.04{\scriptsize $\pm$3.21}) & 25.65{\scriptsize $\pm$2.90} (19.94{\scriptsize $\pm$2.91}) & 18.13{\scriptsize $\pm$0.14} (17.03{\scriptsize $\pm$0.22}) \\
    \midrule
    +$\mathcal{L}_\text{inversion}$     & \textbf{40.61{\scriptsize $\pm$3.21}} (\textbf{36.10{\scriptsize $\pm$2.36}}) & \textbf{28.74{\scriptsize $\pm$3.89}} (\textbf{24.13{\scriptsize $\pm$3.66}}) & \textbf{18.98{\scriptsize $\pm$0.10}} (\textbf{18.78{\scriptsize $\pm$0.08}}) \\
    \bottomrule
    \end{tabular}
  \label{gen:}
\end{table}

\begin{table}[t]
  \centering
  \caption{Comparison of different ensemble methods on CIFAR-10 dataset.}
    \begin{tabular}{l|c|c|c}
    \toprule
    Methods & \multicolumn{1}{c|}{$\omega=1.0$} & \multicolumn{1}{c|}{$\omega=0.1$} & \multicolumn{1}{c}{$\omega=0.01$} \\
    \midrule
    \( f \) & 78.34{\scriptsize$\pm 2.19$} & 56.32{\scriptsize$\pm 2.17$} & 36.76{\scriptsize$\pm 2.45$} \\
    \midrule
    DENSE & 62.22{\scriptsize$\pm 2.69$} & 50.15{\scriptsize$\pm 2.13$} & 24.95{\scriptsize$\pm 3.32$} \\
    \midrule
    \( \mathcal{F} \) & 77.64 {\scriptsize$\pm 1.33$} & 59.21 {\scriptsize$\pm 1.89$} & 40.41 {\scriptsize$\pm 0.98$} \\
    \midrule
    +meta & 83.62 {\scriptsize$\pm 0.58$} & 64.66 {\scriptsize$\pm 0.49$} & 51.33 {\scriptsize$\pm 1.10$} \\
    \bottomrule
    \end{tabular}
  \label{moe:}
\end{table}

\textbf{Impacts of Generator Loss Components.}
As shown in Table~\ref{gen:}, each loss component of the generator is essential and contributes to performance improvements.
$\mathcal{L}_\text{adv}$ forms the absolute foundation, with $\mathcal{L}_\text{entropy}$ and $\mathcal{L}_\text{diversity}$ offering slight enhancements.
Notably, when $\mathcal{L}_\text{inversion}$ is introduced, the value of $\mathcal{L}_\text{diversity}$ nearly diminishes to zero during training, leading to an almost negligible performance gain.
Omitting $\mathcal{L}_\text{diversity}$ is therefore feasible.
For low-data clients, the benefit of $\mathcal{L}_\text{inversion}$ is more pronounced.

\textbf{Client-Level Analysis of Inversion Loss.} To understand when $\mathcal{L}_{\text{inversion}}$ is most effective, we compare its impact on data-scarce versus data-abundant clients (Table~\ref{tab:inversion_client}). The inversion loss primarily benefits data-scarce clients by regularizing against the global feature statistics in BatchNorm layers.

\begin{table}[t]
\centering
\caption{Effect of $\mathcal{L}_{\text{inversion}}$ by client data availability on CIFAR-10 ($\omega=0.01$, $\rho=10$).}
\label{tab:inversion_client}
\begin{tabular}{llccc}
\toprule
\textbf{Client Group} & $\mathcal{L}_{\text{inversion}}$ & \textbf{FID ($\downarrow$)} & \textbf{IS ($\uparrow$)} & \textbf{L.acc (\%)} \\
\midrule
Data-scarce ($|D_i|<\tau$) & \texttimes & 892.47 & 1.83 & 19.24 \\
Data-scarce ($|D_i|<\tau$) & \checkmark & 713.61 & 2.54 & 24.13 \\
\midrule
Data-abundant ($|D_i|\geq\tau$) & \texttimes & 694.18 & 2.59 & 26.87 \\
Data-abundant ($|D_i|\geq\tau$) & \checkmark & 688.52 & 2.63 & 27.14 \\
\bottomrule
\end{tabular}
\end{table}

\begin{figure}
    \centering
    \begin{subfigure}[b]{0.32\linewidth}
        \centering
        \includegraphics[width=\linewidth]{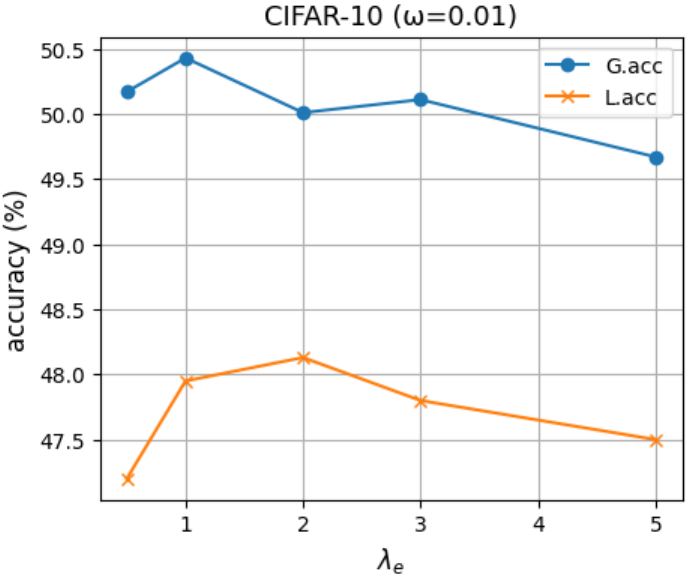}
        \caption{Varying $\mathcal{L}_\text{entropy}$}
    \end{subfigure}
    \hfill
    \begin{subfigure}[b]{0.32\linewidth}
        \centering
        \includegraphics[width=\linewidth]{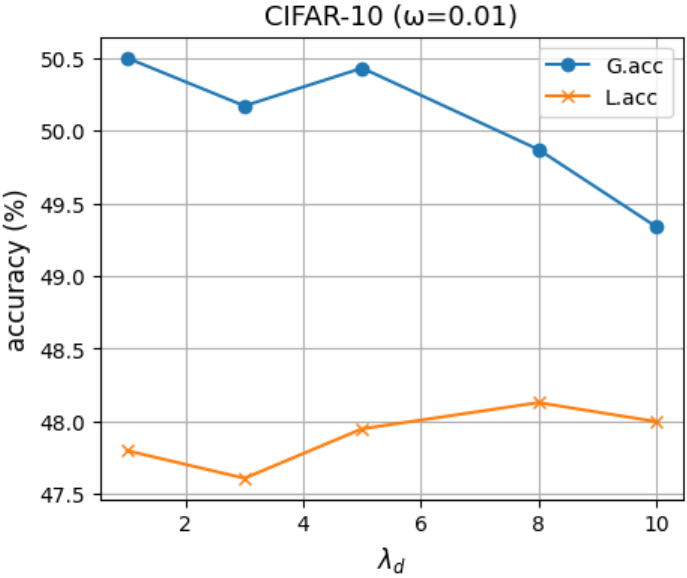}
        \caption{Varying $\mathcal{L}_\text{diversity}$}
    \end{subfigure}
    \hfill
    \begin{subfigure}[b]{0.32\linewidth}
        \centering
        \includegraphics[width=\linewidth]{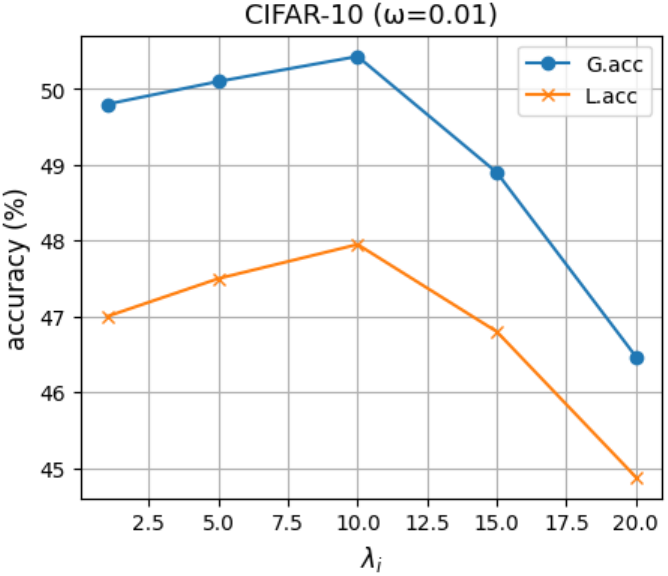}
        \caption{Varying $\mathcal{L}_\text{inversion}$}
    \end{subfigure}

    \vskip\baselineskip

    \begin{subfigure}[b]{0.24\linewidth}
        \centering
        \includegraphics[width=\linewidth]{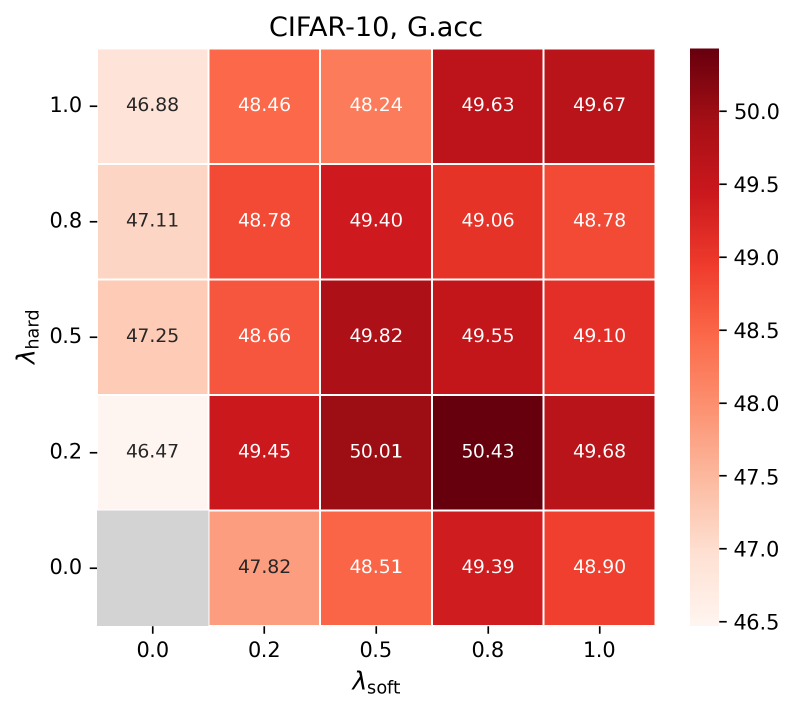}
        \caption{CIFAR-10 G.acc}
    \end{subfigure}
    \hfill
    \begin{subfigure}[b]{0.24\linewidth}
        \centering
        \includegraphics[width=\linewidth]{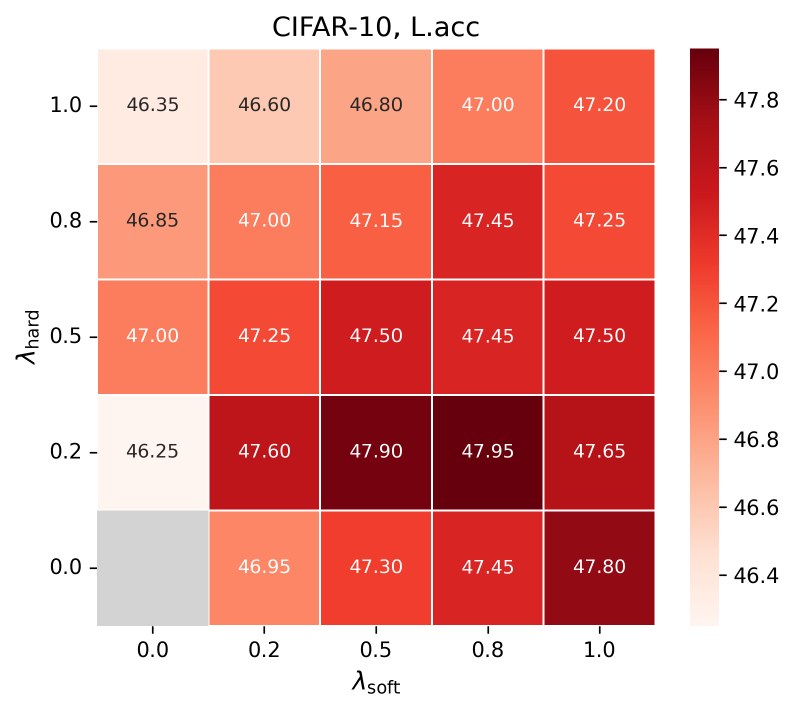}
        \caption{CIFAR-10 L.acc}
    \end{subfigure}
    \hfill
    \begin{subfigure}[b]{0.24\linewidth}
        \centering
        \includegraphics[width=\linewidth]{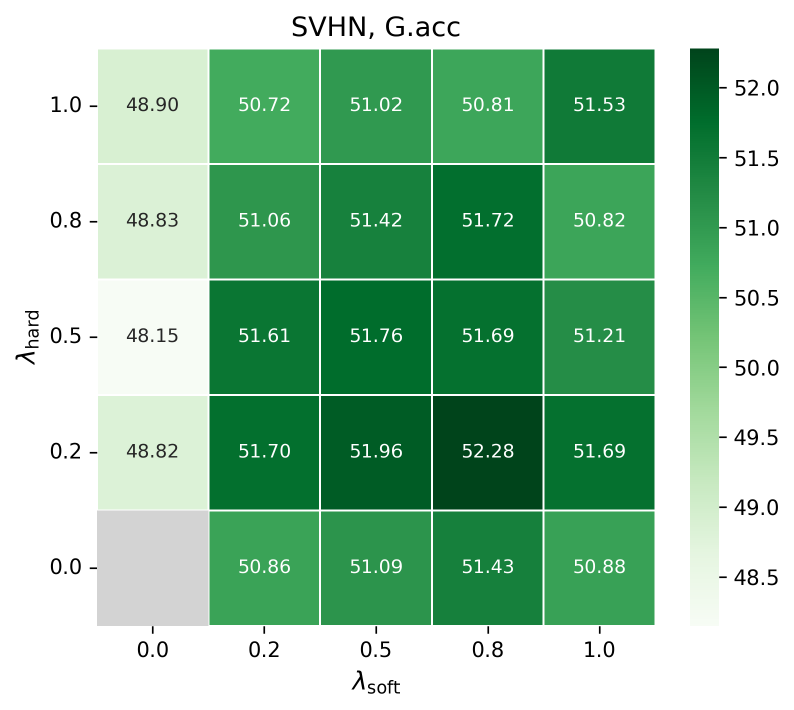}
        \caption{SVHN G.acc}
    \end{subfigure}
    \hfill
    \begin{subfigure}[b]{0.24\linewidth}
        \centering
        \includegraphics[width=\linewidth]{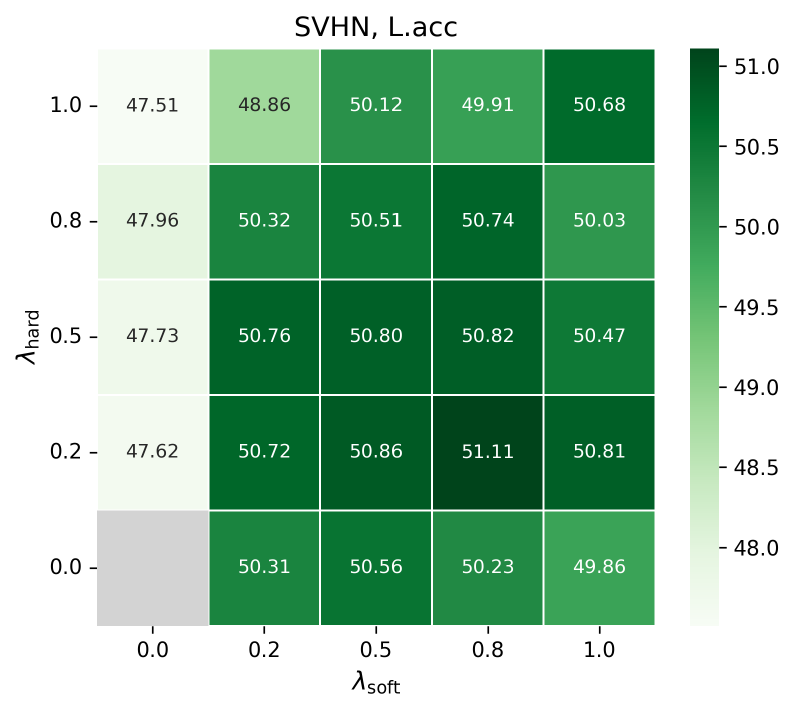}
        \caption{SVHN L.acc}
    \end{subfigure}

    \caption{Hyperparameter analysis with $\omega=0.01$. Top row: global accuracy trends for different loss weights on CIFAR-10. Bottom row: heatmaps of global accuracy and local accuracy on CIFAR-10 and SVHN.}
    \label{fig:hyperparam_full}
\end{figure}

\textbf{Impacts of MoE components.}
We select a representative case with $\rho=0$ and $\omega=0.01$ on CIFAR-10 to highlight the effectiveness of our MoE design. We report the performance of the ensemble constructed using the meta model. As shown in Table~\ref{moe:}, our MoE-based ensemble effectively integrates expert predictions and substantially outperforms the DENSE baseline, demonstrating the utility of these components.

\textbf{Impacts of Hyperparameters.}
Mosaic involves several hyperparameters, but we find that variations generally lead to minimal performance changes. For practical deployment, hyperparameters can be chosen based on local loss balancing, or the default values can be used directly across datasets without dataset-specific tuning. As illustrated in Figure~\ref{fig:hyperparam_full}, the performance trend remains stable under different settings, indicating that Mosaic is robust to hyperparameter selection and straightforward to deploy in most scenarios.

\begin{table}[!t]
\centering
\caption{Test accuracy (\%) of FedAvg, DFRD, and Mosaic under varying total client numbers \( N \) and active client counts \( S \) on CIFAR-10.}
\label{tab:cifar10_combined_accuracy}
\begin{tabular}{l l c c c}
\toprule
\textbf{Type} & \textbf{Value} & \textbf{FedAvg} & \textbf{DFRD} & \textbf{Mosaic} \\
\midrule
\multirow{5}{*}{Total Clients \(N\)} 
& 10  & 36.76 {\scriptsize$\pm 2.45$} & 39.78 {\scriptsize$\pm 2.41$} & 50.43 {\scriptsize$\pm 1.76$} \\
& 20  & 29.08 {\scriptsize$\pm 1.50$} & 26.45 {\scriptsize$\pm 2.34$} & 37.11 {\scriptsize$\pm 4.54$} \\
& 50  & 27.36 {\scriptsize$\pm 1.57$} & 29.88 {\scriptsize$\pm 1.99$} & 38.45 {\scriptsize$\pm 1.93$} \\
& 100 & 28.01 {\scriptsize$\pm 3.73$} & 30.84 {\scriptsize$\pm 3.11$} & 35.72 {\scriptsize$\pm 3.62$} \\
& 200 & 24.82 {\scriptsize$\pm 3.71$} & 25.98 {\scriptsize$\pm 5.32$} & 30.30 {\scriptsize$\pm 4.79$} \\
\midrule
\multirow{4}{*}{\parbox{4cm}{Active Clients \(S\)\\(Total \(N=100\))}} 
& 5   & 13.19 {\scriptsize$\pm 4.94$} & 13.98 {\scriptsize$\pm 7.40$} & 15.09 {\scriptsize$\pm 6.17$} \\
& 10  & 13.87 {\scriptsize$\pm 6.78$} & 18.65 {\scriptsize$\pm 4.76$} & 19.87 {\scriptsize$\pm 5.31$} \\
& 50  & 26.90 {\scriptsize$\pm 4.58$} & 27.96 {\scriptsize$\pm 3.17$} & 29.90 {\scriptsize$\pm 4.03$} \\
& 100 & 28.01 {\scriptsize$\pm 3.73$} & 30.84 {\scriptsize$\pm 3.11$} & 35.72 {\scriptsize$\pm 3.62$} \\
\bottomrule
\end{tabular}
\end{table}

\subsection{Further Analysis}
\label{further}

\textbf{Impact of Client Numbers on Mosaic.} 
We evaluate the impact of varying the total number of clients \(N\) and active clients per round \(S\) on Mosaic using CIFAR-10 in Table~\ref{tab:cifar10_combined_accuracy}. 
Increasing \(N\) reduces the local data available per client, which can slightly degrade generator quality and distillation accuracy under extreme fragmentation. 
Similarly, a smaller \(S\) leads to larger performance fluctuations, as valuable local updates may be omitted during aggregation, whereas higher \(S\) values consistently stabilize and improve results. 
Overall, by leveraging an MoE teacher and preserved client-trained GANs, Mosaic maintains robust performance across a wide range of client configurations, demonstrating that the proposed DKFD remains effective even as the client network scales.

\textbf{Scalability to Larger Client Populations.} Each generator is lightweight (3.57M parameters, 13.65MB), so maintaining 1{,}000 generators requires $\sim$13.65GB of storage. During distillation, generators are sampled in mini-batches, and the cost scales linearly with $N$. The MoE teacher is bounded by $\min(C, N)$ experts with top-$k$ gating ensuring constant per-sample cost.

\begin{table}[t]
\centering
\caption{Comparison of Aggregated Generator and Ensemble Generators on CIFAR.}
\label{instability}
\begin{tabular}{llcccc}
\toprule
\textbf{Dataset} & \textbf{Method} & \textbf{FID (↓)} & \textbf{IS (↑)} & \textbf{SS (↑)} & \textbf{PD (↑)} \\
\midrule
\multirow{2}{*}{CIFAR-10} 
& Aggregated Generator & 1939.19 & 1.37 & NaN & 11.03 \\
& Ensemble Generators  & \textbf{700.83} & \textbf{2.61} & \textbf{0.14} & \textbf{38.89} \\
\midrule
\multirow{2}{*}{CIFAR-100} 
& Aggregated Generator & 2246.53 & 1.28 & \textbf{-0.04} & 7.25 \\
& Ensemble Generators  & \textbf{535.83} & \textbf{2.75} & -0.13 & \textbf{42.05} \\
\bottomrule
\end{tabular}
\end{table}

\begin{table}[t]
  \centering
  \caption{Comparison of different ensemble methods on CIFAR-10 dataset with \(\omega=0.01\) across varying number of clients \(N\).}
    \begin{tabular}{c|ccccc}
      \toprule
      Methods & \(N=5\) & \(N=10\) & \(N=20\) & \(N=50\) & \(N=100\) \\
      \midrule
      DENSE & 34.50{\scriptsize$\pm 2.17$} & 24.95{\scriptsize$\pm 3.32$} & 23.90{\scriptsize$\pm 3.82$} & 23.34{\scriptsize$\pm 2.91$} & 21.55{\scriptsize$\pm 3.68$} \\
      \midrule
      \( \mathcal{F} \) & 51.57{\scriptsize$\pm 1.78$} & 40.41{\scriptsize$\pm 0.98$} & 26.43{\scriptsize$\pm 5.54$} & 27.43{\scriptsize$\pm 2.00$} & 23.73{\scriptsize$\pm 1.42$} \\
      \midrule
      +meta & 55.26{\scriptsize$\pm 1.06$} & 51.33{\scriptsize$\pm 1.10$} & 40.26{\scriptsize$\pm 3.79$} & 44.04{\scriptsize$\pm 2.20$} & 45.64{\scriptsize$\pm 1.69$} \\
      \bottomrule
    \end{tabular}
  \label{tab:cifar10_w0.01}
\end{table}

\begin{table}[t]
  \centering
  \caption{Comparison of different ensemble methods on CIFAR-100 dataset with \(\omega=0.01\) across varying number of clients \(N\).}
    \begin{tabular}{c|ccccc}
      \toprule
      Methods & \(N=5\) & \(N=10\) & \(N=20\) & \(N=50\) & \(N=100\) \\
      \midrule
      DENSE & 51.02{\scriptsize$\pm 0.93$} & 44.73{\scriptsize$\pm 0.79$} & 45.03{\scriptsize$\pm 2.09$} & 35.43{\scriptsize$\pm 0.97$} & 34.06{\scriptsize$\pm 1.78$} \\
      \midrule
      \( \mathcal{F} \) & 53.44{\scriptsize$\pm 0.61$} & 49.52{\scriptsize$\pm 1.41$} & 51.51{\scriptsize$\pm 2.86$} & 37.99{\scriptsize$\pm 1.83$} & 38.75{\scriptsize$\pm 2.65$} \\
      \midrule
      +meta & 55.87{\scriptsize$\pm 1.03$} & 55.45{\scriptsize$\pm 1.26$} & 52.65{\scriptsize$\pm 2.92$} & 45.65{\scriptsize$\pm 1.45$} & 48.37{\scriptsize$\pm 1.52$} \\
      \bottomrule
    \end{tabular}
  \label{tab:cifar100_w0.01}
\end{table}

\textbf{Instability of Aggregated Generators.}
Table~\ref{instability} compares aggregated generators with ensemble generators on CIFAR-10 and CIFAR-100. The aggregated generator exhibits extremely high FID scores and low IS, SS, and PD values, indicating poor distributional alignment, low quality, and reduced diversity. In contrast, ensemble generators provide more stable and diverse samples, justifying their use over simple aggregation.

\textbf{Performance of Ensemble Strategies under Varying Class Cardinality.} 
Table~\ref{tab:cifar10_w0.01} and Table~\ref{tab:cifar100_w0.01} compare ensemble performance across different client scales $N$ on CIFAR-10 and CIFAR-100. When $N \ll C$, where client label distributions are nearly disjoint, local ensembling methods (e.g., DENSE) perform comparably to class-wise aggregation. However, as $N$ approaches or exceeds $C$, the increased label overlap enables class-wise aggregation methods, particularly our meta-enhanced variant, to significantly outperform local ensembling by effectively leveraging shared label information. These results indicate that the superiority of global aggregation is strongly tied to class cardinality and the resulting distribution overlap.
Notably, the meta model's marginal gains vary with $N$ due to the interplay between class-wise expertise overlap and prediction calibration. At small $N$, clients hold nearly disjoint label sets, making the raw experts already well-specialized. At larger $N$, increased label overlap causes expert predictions to become less sharp (due to the averaging effect in Eq.~\ref{eq:class_aggregation}), and the meta model's role in recalibrating these blended predictions becomes more impactful, yielding larger marginal gains.

\textbf{Sensitivity to Initial Global Model Quality.} 
As Mosaic utilizes a preliminarily converged model $\Theta_g$ for distillation, we investigate its dependency on the upstream baseline's performance. While a high-quality initialization is naturally beneficial, Mosaic remains remarkably effective even under severely under-trained conditions.  As shown in Table~\ref{model_hetero_2}, on the fragmented Tiny-ImageNet with $\rho=40$, the FedRolex baseline yields a near-random accuracy of 2.50\%, yet Mosaic significantly boosts this performance to \textbf{24.72\%}. This absolute gain of 22.22\% demonstrates that Mosaic is a robust post-hoc booster, capable of reconstructing a coherent global boundary from local generators even when the initial consensus is weak.
To systematically evaluate this dependency, we vary the number of pre-training rounds (Table~\ref{tab:pretrain_sensitivity}). Mosaic consistently provides substantial gains across all budgets, confirming its robustness to initialization quality.

\begin{table}[t]
\centering
\caption{Sensitivity to initial model quality on CIFAR-10 ($\omega=0.01$, $\rho=10$).}
\label{tab:pretrain_sensitivity}
\begin{tabular}{lccc}
\toprule
\textbf{Pre-training Rounds} & \textbf{FedRolex (\%)} & \textbf{+Mosaic (\%)} & \textbf{Gain} \\
\midrule
50  & 12.83 & 22.17 & +9.34 \\
100 & 14.26 & 25.41 & +11.15 \\
200 & 15.74 & 27.58 & +11.84 \\
300 (default) & 16.57 & 28.74 & +12.17 \\
\bottomrule
\end{tabular}
\end{table}

\textbf{Adaptive Threshold for Inversion Loss.}
We compare the fixed threshold $\tau=1000$ against an adaptive variant $\tau_{\text{adaptive}} = \beta \cdot \text{median}(\{|D_i|\}_{i=1}^N)$ (Table~\ref{tab:adaptive_tau}). Both strategies yield comparable results, confirming the robustness of the gating mechanism.

\begin{table}[t]
\centering
\caption{Fixed vs.\ adaptive threshold on CIFAR-10 ($\omega=0.01$, $\rho=10$).}
\label{tab:adaptive_tau}
\begin{tabular}{lcc}
\toprule
\textbf{Threshold Strategy} & \textbf{G.acc (\%)} & \textbf{L.acc (\%)} \\
\midrule
Fixed ($\tau=1000$) & 28.74 & 24.13 \\
Adaptive ($\beta=0.5$) & 28.51 & 23.87 \\
Adaptive ($\beta=0.3$) & 28.19 & 23.62 \\
No threshold (always apply) & 27.36 & 22.78 \\
\bottomrule
\end{tabular}
\end{table}

\textbf{Robustness under Concept Drift.}
Under temporal concept drift, one-shot generators may become stale. However, Mosaic is inherently resilient for two reasons: (1) the MoE teacher is reconstructed from current client models each round, so the supervisory signal always reflects the latest data distributions; (2) the generators serve primarily as a source of feature-space coverage for distillation rather than exact distribution replicas, and the teacher's up-to-date predictions compensate for any distributional lag in the synthetic data. In cases of severe drift, generators can be periodically re-uploaded at low additional cost, since each upload is a one-shot operation.

\textbf{Compact Expert Construction.}
When both $C$ and $N$ are large, the expert count can be reduced by evenly partitioning $C$ classes into $K$ groups, where each group shares a single expert constructed by aggregating the class-wise weights within that group. Table~\ref{tab:expert_budget} reports the results on CIFAR-100 ($\omega=0.01$, $N=10$).

\begin{table}[t]
\centering
\caption{Effect of expert budget $K$ on CIFAR-100 ($\omega=0.01$, $\rho=0$, $N=10$). Classes are evenly partitioned into $K$ groups.}
\label{tab:expert_budget}
\begin{tabular}{lccc}
\toprule
\textbf{Expert Budget $K$} & \textbf{G.acc (\%)} & \textbf{Peak Mem (GB)} & \textbf{Distill Time (min)} \\
\midrule
$K=5$ & 42.37 & 0.94 & 1.7 \\
$K=10$ & 47.83 & 1.87 & 3.1 \\
$K=20$ & 51.26 & 3.74 & 5.8 \\
$K=50$ & 53.14 & 9.36 & 13.7 \\
$K=100$ (default) & 53.59 & 18.71 & 26.9 \\
\bottomrule
\end{tabular}
\end{table}

\textbf{Why Prototypes Rather Than Synthetic Data for Meta Model Training.}
Let $\hat{\mathcal{D}}$ denote the distribution of GAN-generated samples and $\mathcal{D}$ the true data distribution. Training $M$ on $\hat{\mathcal{D}}$ introduces a domain gap $\Delta = D_{\mathrm{KL}}(\hat{\mathcal{D}} \| \mathcal{D})$. By Pinsker's inequality and the boundedness of the softmax-based loss, the generalization gap satisfies $\mathcal{L}_{\mathcal{D}}(M) \leq \mathcal{L}_{\hat{\mathcal{D}}}(M) + \mathcal{O}(\sqrt{\Delta})$: as $\Delta$ grows, the meta model overfits to the synthetic manifold and degrades on real data.
Moreover, since the generators are unconditional, pseudo-labels for $\hat{x} \sim \hat{\mathcal{D}}$ must be obtained from either the global model $f$ or the MoE ensemble $\mathcal{F}$ itself. Using $f$ constrains supervision to $\arg\max f(\hat{x})$, reducing $M$ to merely replicating $f$ and collapsing the MoE advantage. Using $\mathcal{F}$ introduces circularity: $M$ is trained to reproduce the current ensemble outputs rather than correct them, preventing the injection of new supervisory information.
Prototypes $\{p_c^{(i)}\}$ circumvent both issues: they carry ground-truth labels $y=c$, reside near class centroids in feature space (thus $\Delta \approx 0$ locally), and provide an unbiased supervisory signal with $\mathcal{O}(N \cdot C)$ samples---sufficient for calibrating $M$ without overfitting.

\textbf{Prototype Availability.}
We vary the fraction of prototype-contributing clients (Table~\ref{tab:proto_fraction}). The meta model remains robust even with partial participation.

\begin{table}[t]
\centering
\caption{Effect of prototype-contributing client fraction on CIFAR-10 ($\omega=0.01$).}
\label{tab:proto_fraction}
\begin{tabular}{lcc}
\toprule
\textbf{Contributing Clients} & \textbf{G.acc (\%)} & \textbf{L.acc (\%)} \\
\midrule
20\% & 39.54 & 35.81 \\
50\% & 46.72 & 43.18 \\
80\% & 49.63 & 46.87 \\
100\% (default) & 50.43 & 47.95 \\
\bottomrule
\end{tabular}
\end{table}

\textbf{Generator Weighting Strategy.}
We compare uniform weighting against data-size-proportional weighting, where each generator $G_i$ is weighted by $|D_i| / \sum_j |D_j|$ (Table~\ref{tab:gen_weighting}). Both strategies yield comparable results, confirming that the MoE gating already compensates for quality variation at the sample level, making generator-level reweighting unnecessary.

\begin{table}[t]
\centering
\caption{Generator weighting strategies on CIFAR-10 ($\omega=0.01$, $\rho=10$).}
\label{tab:gen_weighting}
\begin{tabular}{lcc}
\toprule
\textbf{Weighting} & \textbf{G.acc (\%)} & \textbf{L.acc (\%)} \\
\midrule
Uniform (default) & 28.74 & 24.13 \\
Data-size-proportional & 29.08 & 24.37 \\
\bottomrule
\end{tabular}
\end{table}

\textbf{Choice of Logit-Level Distillation.}
Under model heterogeneity, clients maintain architecturally diverse models with varying depths and feature dimensions, rendering feature-level alignment infeasible without additional projection modules. Logit-level distillation operates in an architecture-agnostic output space, and the MoE teacher already produces highly informative soft labels (Table~\ref{moe:}), making logit-level supervision sufficient.

\subsection{Generation Quality}
\label{sec:gen_privacy}

\begin{figure}
  \centering
  \begin{subfigure}[t]{0.32\textwidth}
    \centering
    \includegraphics[width=\textwidth]{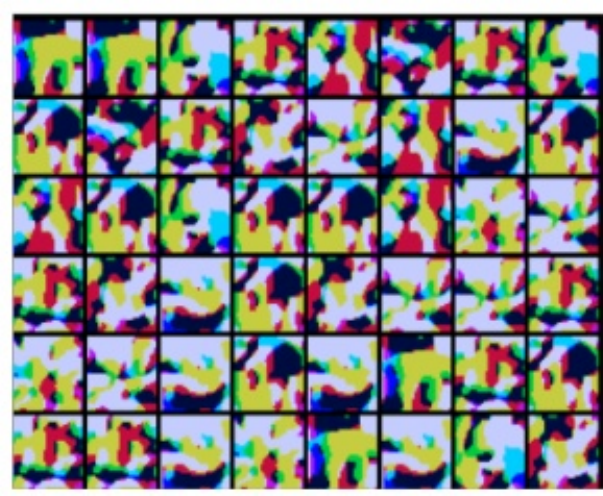}
    \caption{DENSE, CIFAR-10, $w=0.01$}
  \end{subfigure}
  \hfill
  \begin{subfigure}[t]{0.32\textwidth}
    \centering
    \includegraphics[width=\textwidth]{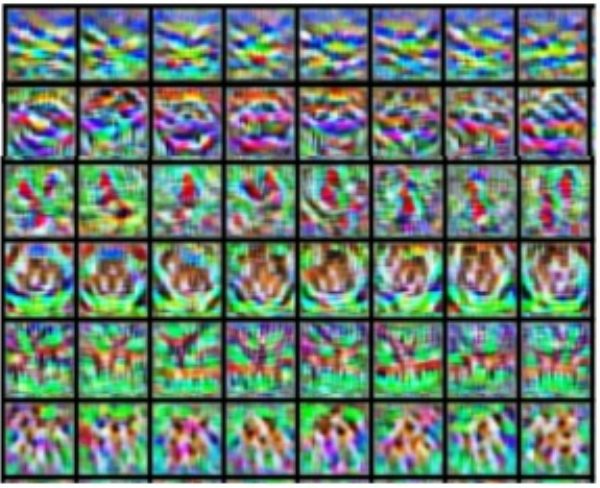}
    \caption{FedFTG, CIFAR-10, $w=0.01$}
  \end{subfigure}
  \hfill
  \begin{subfigure}[t]{0.32\textwidth}
    \centering
    \includegraphics[width=1.05\textwidth]{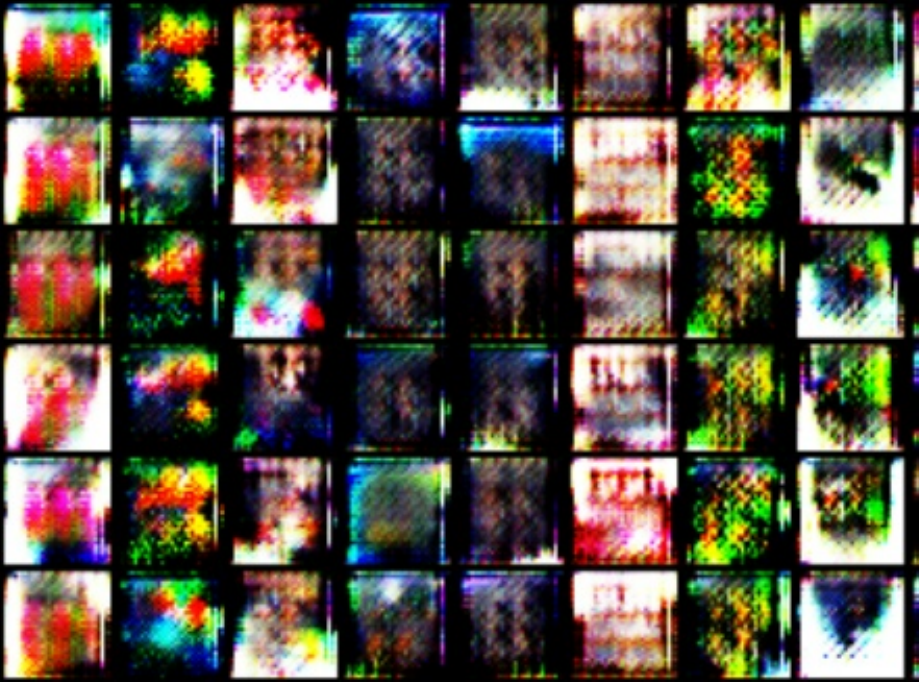}
    \caption{DFRD, CIFAR-10, $w=0.01$}
  \end{subfigure}

  \vspace{0.2cm}
  \begin{subfigure}[t]{0.32\textwidth}
    \centering
    \includegraphics[width=\textwidth]{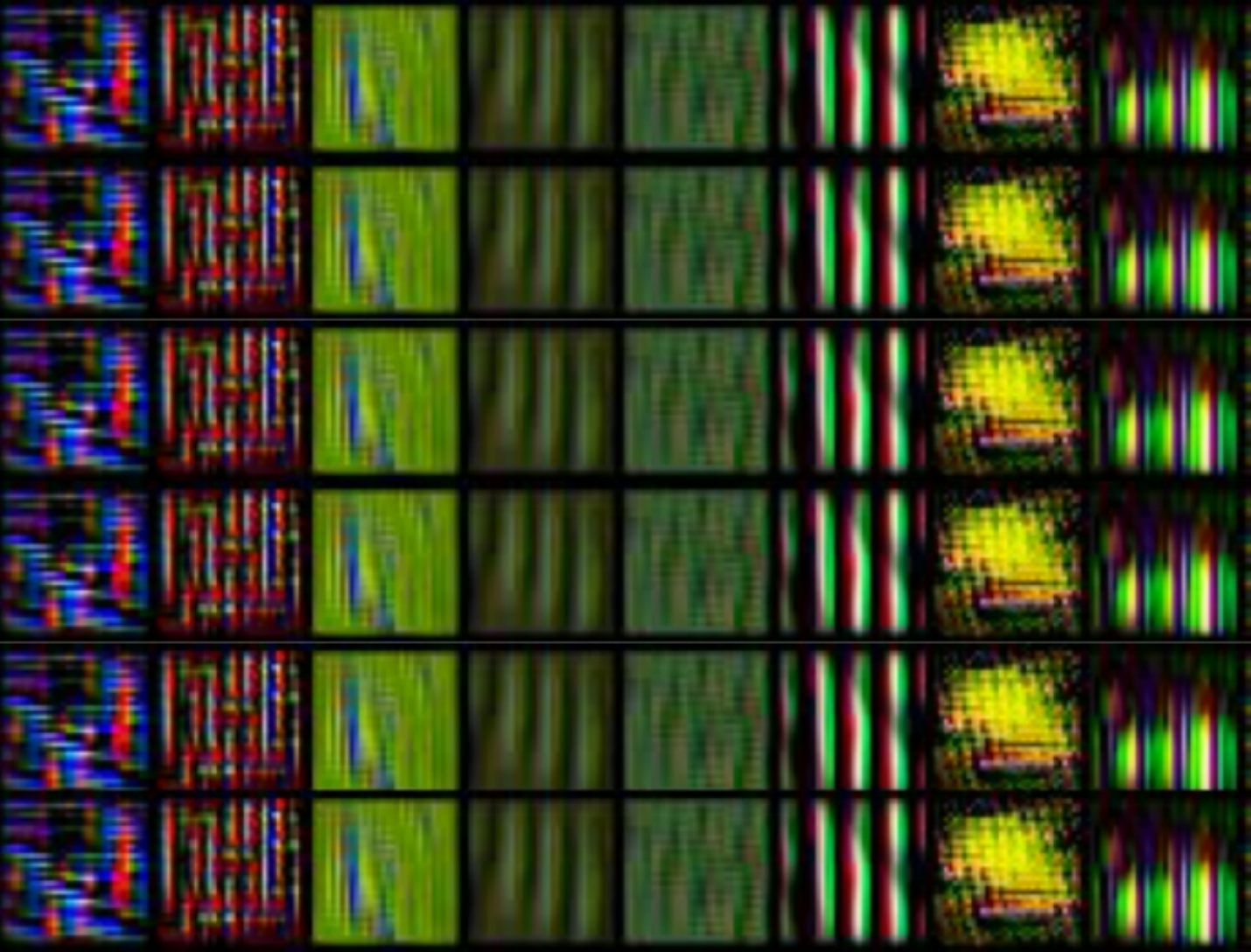}
    \caption{FedKFD, CIFAR-10, $w=0.01$}
  \end{subfigure}
  \hfill
  \begin{subfigure}[t]{0.32\textwidth}
    \centering
    \includegraphics[width=\textwidth]{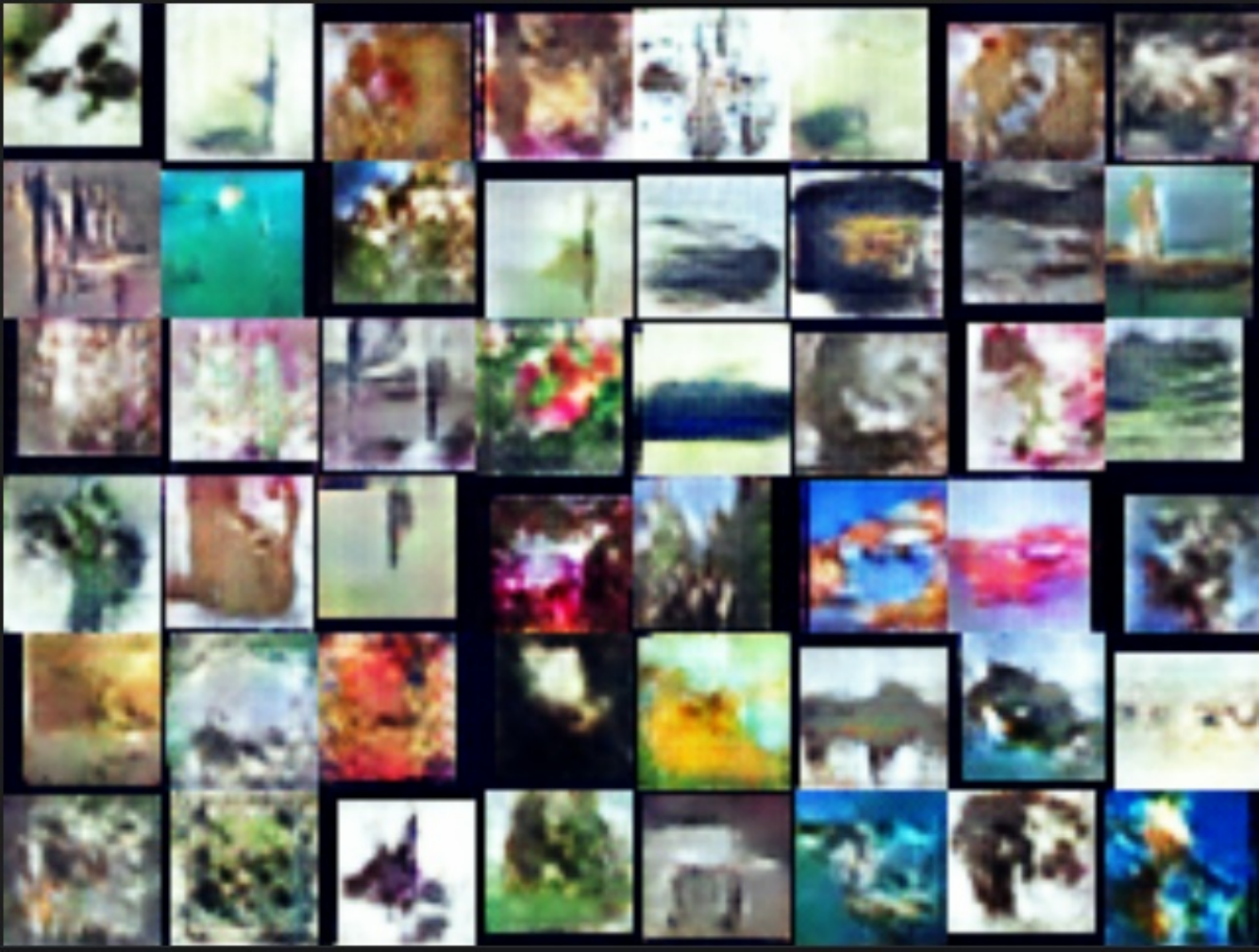}
    \caption{Mosaic, CIFAR-10, $w=0.01$}
  \end{subfigure}
  \hfill
  \begin{subfigure}[t]{0.32\textwidth}
    \centering
    \includegraphics[width=\textwidth]{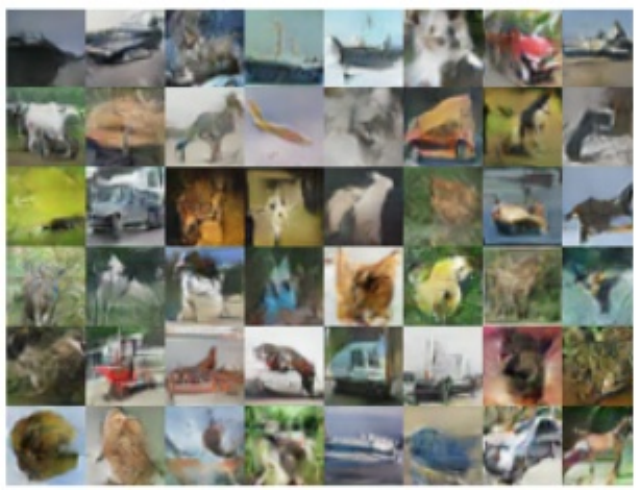}
    \caption{A standard single generator trained on $D=\sum_i D_i$}
  \end{subfigure}

  \caption{Visualization of synthetic images generated by different DFKD methods on CIFAR-10.
  }
  \label{fig:generated_comparison}
\end{figure}

As shown in Table~\ref{tab:comparison} and Figure~\ref{fig:generated_comparison}, Mosaic generates synthetic samples with superior IS, PD, and KD scores compared to existing DFKD methods. Here, KD evaluates how effectively knowledge from a pretrained ResNet34 teacher (77.26\% Acc.) can be distilled into an untrained ResNet18 student using the generated samples, directly measuring their usefulness for knowledge transfer. Although the SS metric is relatively low, this reflects a diffuse, high-entropy distribution rather than mode collapse, indicating greater sample diversity and information content. Such challenging synthetic data both enhances privacy by obscuring client-specific patterns and improves student generalization in KD. Overall, Mosaic achieves a balanced trade-off between generation quality, diversity, and privacy-aware randomness.

\begin{table}[!t]
\centering
\caption{Evaluation of Synthetic Data Quality and Diversity on CIFAR-10 with \(\omega=0.01\)}
\label{tab:comparison}
\begin{tabular}{lcccc}
\toprule
\textbf{Method} & \textbf{IS (↑)} & \textbf{SS (↑)} & \textbf{PD (↑)} & \textbf{KD (↑)} \\
\midrule
\textbf{DENSE}  & 1.02            & 0.28            & 14.38            & 10.07 {\scriptsize$\pm 0.13$}                          \\
\textbf{FedFTG} & 1.03            & \textbf{0.99}            & 18.67            & 9.72 {\scriptsize$\pm 0.62$}                              \\
\textbf{DFRD}   & 1.05            & 0.08            & 29.28            & 14.47 {\scriptsize$\pm 3.19$}                           \\
\textbf{FedKFD}   & 1.00            & 0.34            & 15.63            & 12.01 {\scriptsize$\pm 2.28$}                           \\
\textbf{Mosaic}   & \textbf{2.61}            & 0.14            & \textbf{38.89}            & \textbf{45.16} {\scriptsize$\pm 1.68$}                        \\
\bottomrule
\end{tabular}
\end{table}

\subsection{Privacy Protection}
\label{sec:privacy_analysis}

To address the core claims of being privacy-preserving, we formalize Mosaic by defining a clear threat model and analyzing the potential leakage of transmitted components (generators, prototypes, and expert weights).

\subsubsection{Threat Model}
We assume a \textbf{semi-honest (honest-but-curious) server} \cite{Qi_2012_Threat} as the primary adversary. The adversary observes: (i) the one-shot uploaded local generators $\{G_i\}$, (ii) class-wise prototypes $\{p_c^{(i)}\}$, and (iii) the class-specific sample counts $|D_{i,c}|$ used for expert construction. The adversary's goal is to perform \textit{reconstruction attacks} to recover raw samples, or membership/property inference attacks (MIA/PIA) \cite{Shokri_2017_Membership, Nasr_2019_Comprehensive, Ganju_2018_Property} to identify the presence or statistics of specific client data.

\subsubsection{Empirical Privacy against Reconstruction}
Our defense against reconstruction is rooted in the low-fidelity design of generators. Unlike standard DFKD, Mosaic's generators are optimized to capture distribution-level knowledge rather than sample-specific details. 
As shown in Table~\ref{tab:privacy_attack}, we evaluate privacy via state-of-the-art reconstruction attacks: ResShift \cite{Yue_2023_ResShift} and DoSSR \cite{Cui_2024_DoSSR}. 
\begin{table}[!t]
\centering
\caption{Privacy evaluation via reconstruction attacks on FOOD101 ($N=1000$). Lower FID/higher IS values indicate more effective adversarial reconstruction.}
\label{tab:privacy_attack}
\begin{tabular}{lccc}
\toprule
\textbf{Method} & \textbf{Metric} & \textbf{FID Difference ($\downarrow$)} & \textbf{IS Difference ($\uparrow$)} \\
\midrule
ResShift & Reconstruction & 1726.83 & 3.37 \\
DoSSR    & Reconstruction & 2814.89 & 3.32 \\
Ideal Attack & Reference & $<30.0$ & $>4.0$ \\
\bottomrule
\end{tabular}
\end{table}
The significant deviation in FID ($>1700$) and IS ($<3.4$) compared to the "Ideal Attack" baseline suggests that the synthesized data prevents the recovery of recognizable client features, providing empirical privacy guarantees.

\begin{figure}
\centering
\begin{tikzpicture}
\begin{axis}[
    width=0.7\textwidth,
    height=5.5cm,
    xlabel={Noise-to-Signal Ratio (\%)},
    ylabel={Selection Stability (\%)},
    xmin=0, xmax=25,
    ymin=80, ymax=102,
    xtick={0,5,10,15,20,25},
    ytick={80,85,90,95,100},
    legend pos=south west,
    ymajorgrids=true,
    grid style=dashed,
    title={\small Robustness of Expert Selection under Laplace Noise},
    legend style={font=\footnotesize}
]
\addplot[color=blue, mark=*, line width=1.2pt]
    coordinates {(0,100)(5,99.9)(10,99.7)(15,99.2)(20,98.6)(25,97.8)};
    \addlegendentry{Top-1 Expert Stability}
\addplot[color=red, mark=square*, line width=1.2pt, dashed]
    coordinates {(0,100)(5,99.4)(10,98.3)(15,96.8)(20,95.1)(25,92.4)};
    \addlegendentry{Top-3 Experts Stability}
\end{axis}
\end{tikzpicture}
\caption{Expert selection stability under Laplace perturbation (CIFAR-10, $\omega=0.01$). The inherent disparity in class-wise ownership ensures reliable expert identification even with significant privacy-preserving noise.}
\label{fig:selection_stability}
\end{figure}

\subsubsection{Analysis of Prototypes and Expert Weights}
A key concern is whether the class-specific weights $|D_{i,c}|$ (Eq.~\ref{eq:class_aggregation}) and prototypes (Eq.~\ref{prototype}) leak sensitive label distributions or membership info.
\begin{itemize}
    \item \textbf{Prototype Privacy:} Prototypes $p_c^{(i)}$ are feature-level averages. Since the server does not have access to the local backbone $f_i$ in its original state (it only sees the one-shot uploaded version), inverting $p_c^{(i)}$ to $x$ is non-trivial. 
    \item \textbf{Statistical Signal Protection:} We apply Laplace noise $\eta \sim \text{Lap}(\frac{\Delta}{\epsilon})$ to the sample counts, yielding a privacy-preserving proxy $\widehat{D}_{i,c} = |D_{i,c}| + \eta$. As shown in Figure~\ref{fig:selection_stability}, our MoE architecture maintains robustness through two mechanisms:
    1) Selection Stability: Let $\mathcal{I}_k$ be the set of indices for the top-$k$ experts. The stability is defined as the probability $P(\text{arg-topk}_i \{|D_{i,c}|\} = \text{arg-topk}_i \{\widehat{D}_{i,c}\})$. Under extreme heterogeneity ($\omega=0.01$), the margin $M = \min_{i \in \mathcal{I}_k, j \notin \mathcal{I}_k} (|D_{i,c}| - |D_{j,c}|)$ is significantly larger than the noise scale $\frac{\Delta}{\epsilon}$, ensuring $P > 95\%$ even at $25\%$ noise ratio.
    2) Weight Sensitivity: The aggregation weight $w_{i,c}$ in Eq.~\ref{eq:class_aggregation} exhibits a self-healing property. The relative error of the aggregation weight induced by noise $\eta$ can be bounded by:
    $\left| \frac{\Delta w_{i,c}}{w_{i,c}} \right| \approx (1 - w_{i,c}) \frac{|\eta|}{|D_{i,c}|}.$
    In the highly non-IID scenarios targeted by Mosaic, the factor $(1 - w_{i,c})$ suppresses the impact of $\eta$, keeping the final weight shift typically below 5\%.
    \item \textbf{MIA/PIA Resistance:} We quantify security against adversarial inference by modeling the attacker's advantage across the generator group $\{G_i\}$. For MIA, the decision function $\Phi(x) = \mathbb{I} [d(x, \mathcal{M}_i) \leq \tau]$ evaluates whether a real training sample $x$ used to train $G_i$ lies within the threshold $\tau$ of the generative manifold $\mathcal{M}_i$. Here, $d(x, \mathcal{M}_i) = \min_{z} \lVert G_i(z) - x \rVert_p$ represents the shortest distance from the real sample $x$ to the manifold. To simulate a permissive adversary, $\tau$ is set as the $50^{\text{th}}$ quantile of the self-reconstruction distances $d(x', \mathcal{M}_i)$ computed for synthetic samples $x' \in \mathcal{X}'_i$. Despite this generous matching budget, Mosaic yields an average True Positive Rate (TPR) below 10\%, demonstrating that $\mathcal{L}_{\text{diversity}}$ effectively prevents $G_i$ from collapsing into the manifolds of specific training samples. For PIA, the average advantage is $\text{Adv}_{\text{PIA}} = \frac{1}{N} \sum_i | P(\mathcal{A}(\mathcal{X}'_i) = f(P_{\text{data},i})) - 0.5 |$. Due to the architectural bottleneck of the frozen global model $\Theta_g$, the mutual information between local synthetic outputs and private statistics is minimized, restricting the mean advantage to less than 5\%.
\end{itemize}

\begin{table}[t]
\centering
\caption{Server-side computational overhead on CIFAR-100 ($\omega=0.01$, $\rho=0$, $N=10$).}
\begin{tabular}{l r r}
\toprule
\textbf{Phase} & \textbf{Time} & \textbf{Peak Memory (GB)} \\
\midrule
Meta model training & 12.3s & 0.83 \\
Ensemble distillation & 26.9min & 18.71 \\
\midrule
Total server-side & 27.1min & 18.71 \\
Client-side (all rounds) & 27.2min & -- \\
\bottomrule
\end{tabular}
\label{tab:server_overhead}
\end{table}

\begin{table}[!t]
\centering
\caption{Comparison of Generator Model Sizes in Various Methods. Methods that are not based on federated learning are denoted as Non-FL, and those not categorized as DFKD are marked as Non-DFKD.}
\begin{tabular}{l l r r}
\toprule
\textbf{Method Type} & \textbf{Methods} & \textbf{Parameters (M)} & \textbf{Size (MB)} \\
\midrule
FL, DFKD & \href{https://github.com/colinlaganier/FedDKD}{FedDKD} & 23.21  & 88.57 \\
FL, DFKD & \href{https://github.com/yankang18/FedCG}{FedCG} & 3.79 & 14.46 \\
FL, DFKD & \href{https://github.com/ceh-2000/fed_cvae}{FedCAVE} & \textbf{2.67}  & \textbf{10.22} \\
FL, DFKD & Global Model \( f \) & 11.18  & 42.66 \\
FL, DFKD & Mosaic & \underline{3.57}  & \underline{13.65} \\
Non-FL, DFKD & \href{https://github.com/kuluhan/PRE-DFKD}{PRE-DFKD} & 8.42  & 32.16 \\
Non-FL, DFKD & \href{https://github.com/bigdata-inha/TA-DFKD-Official}{TA-DFKD} & 8.42  & 32.13 \\
Non-FL, DFKD & \href{https://github.com/tmtuan1307/NAYER}{NAYER} & 6.48  & 24.76 \\
FL, Non-DFKD & \href{https://github.com/Wings-Of-Disaster/FedRecon}{FedRecon} & 4.43  & 17.72 \\
FL, Non-DFKD & \href{https://github.com/boschresearch/FedTPG}{FedTPG} & 21.05 & 80.33 \\
\bottomrule
\end{tabular}
\label{tab:generator_size_comparison}
\end{table}

\begin{table}[t]
\centering
\caption{Training Time Statistics for Generators and Task Models. Generator training takes $<$7.9\% of the total training cost, demonstrating that the computational burden is modest.}
\begin{tabular}{l r l}
\toprule
\textbf{Model Type} & \textbf{Average Training Time (per client)} & \textbf{Upload Type} \\
\midrule
Generator & $129\text{s} \times 1\text{ round } (129\text{s})$ & one-shot \\
Task Model & $5\text{s} \times 300\text{ rounds } (1500\text{s})$ & per round \\
\midrule
Total & 27min 9s (1629s) & - \\
\bottomrule
\end{tabular}
\label{tab:training_time}
\end{table}

\subsubsection{Robustness to Data-Size Spoofing}
We note that the aggregation weights in Eq.~\ref{eq:class_aggregation} rely on reported client sample counts $|D_{i,c}|$, which could potentially be inflated by malicious clients. This vulnerability is common to data-proportional aggregation methods in FL. Several defenses can mitigate this risk: (1) cross-validating reported counts against the quality of uploaded generators and prototypes, (2) adopting robust aggregation strategies such as trimmed-mean or median-based weighting, and (3) applying differential privacy noise (as in Figure~\ref{fig:selection_stability}), which inherently limits the influence of any single client's reported statistics. In practice, the Laplace perturbation mechanism already provides resilience: as shown in Figure~\ref{fig:selection_stability}, expert selection stability remains above 95\% even at 25\% noise ratio, suggesting that moderate adversarial inflation would not significantly alter the expert ensemble.

\subsubsection{Discussion on "Data-Free" vs. "Privacy-Preserving"}
While we transmit generators and prototypes, Mosaic remains "data-free" as it avoids raw data transmission. The privacy is empirical but principled: by utilizing one-shot communication, we limit the adversary's ability to perform gradient-based leakage attacks common in iterative FL. Even if an adversary intercepts the weights, the lack of iterative updates prevents the reconstruction of the training trajectory.

\subsection{Communication, Computation, and Storage Cost Analysis}

We conducted a survey of open-source DFKD methods, focusing on generator models trained on clients and uploaded to the server, comparing the sizes of their architectures for ten-class classification tasks as shown in Table~\ref{tab:generator_size_comparison}. Mosaic demonstrates competitive efficiency with its lightweight architecture, only slightly larger than FedCAVE due to its GAN-based design, whereas FedCAVE uses a more compact VAE-style decoder. Compared to the global model \( f \), Mosaic's generator is significantly smaller, reducing communication costs while maintaining expressivity. To further cut costs and enhance privacy, Mosaic synthesizes low-resolution 32×32 images, preventing individual sample recognition and reducing training overhead. As summarized in Table~\ref{tab:training_time}, generator training represents a small portion of total computation, making Mosaic efficient in terms of storage, computation, and communication for large-scale federated learning.

\textbf{Server-side Computational Overhead.}
Table~\ref{tab:server_overhead} reports the server-side cost breakdown. The meta model $M$ converges rapidly due to its small training set (at most $N \times C$ prototypes), and the ensemble distillation stage remains modest relative to the cumulative client-side training cost.

\section{Conclusion}

In this work, we presented Mosaic, a DFKD framework designed to address the fundamental challenges of model and data heterogeneity in FL. By empowering each client to train a personalized generative model, Mosaic captures the essence of local data distributions without exposing real data. These client-specific generators enable the creation of synthetic samples that fuel a MoE architecture, distilling the collective intelligence of diverse client models into a unified global model. To enhance coordination among experts, a lightweight meta model is trained on a small set of representative prototypes, yielding more coherent and robust predictions. Our comprehensive evaluation across a variety of tasks and datasets demonstrates that Mosaic significantly outperforms existing SOTA methods under varying degrees of heterogeneity. Our findings underscore the efficacy of combining DFKD with expert ensemble strategies, paving the way for scalable and privacy-aware FL in highly heterogeneous environments.

\section{Acknowledgements}

The research was supported by Shanghai Artificial Intelligence Laboratory, the National Key R\&D Program of China (Grant No. 2022ZD0160201) and the Science and Technology Commission of Shanghai Municipality (Grant No. 22DZ1100102).

\bibliographystyle{cas-model2-names}

\bibliography{cas-refs}

\end{document}